\documentclass{ieeeaccess}
\usepackage{cite}
\usepackage{amsmath,amssymb,amsfonts}
\usepackage{algorithmic}
\usepackage{graphicx}
\usepackage{textcomp}

\usepackage{amssymb} 
\usepackage{float} 
\usepackage{gensymb} 
\usepackage[colorlinks=false, pdfborder={0 0 0}]{hyperref} 
\usepackage{multirow} 
\usepackage{rotating} 
\usepackage{soul} 
\usepackage{tabularx} 
\usepackage{threeparttable} 
\usepackage[table]{xcolor} 
\usepackage{ulem}  

\definecolor{my_DarkGreen}{RGB}{0, 128, 0}
\definecolor{MyDarkGreen}{RGB}{0, 100, 0}
\definecolor{MyGoldenrod}{RGB}{218, 165, 32}
\definecolor{MyCrimson}{RGB}{220, 20, 60}

\def\BibTeX{{\rm B\kern-.05em{\sc i\kern-.025em b}\kern-.08em
    T\kern-.1667em\lower.7ex\hbox{E}\kern-.125emX}}

\begin{document}
\history{Date of publication xxxx 00, 0000, date of current version xxxx 00, 0000.}
\doi{10.1109/ACCESS.2023.0322000}

\title{SLAM for Visually Impaired People: a Survey}

\author{Marziyeh Bamdad\authorrefmark{1,2}, Davide Scaramuzza\authorrefmark{1}, and Alireza Darvishy\authorrefmark{2}}

\address[1]{Department of Informatics, University of Zurich, 8050 Zurich, Switzerland}
\address[2]{Institute of Applied Information Technology, Zurich University of Applied Sciences, 88400 Winterthur, Switzerland}

\markboth
{Bamdad \headeretal: SLAM for Visually Impaired People: a Survey}
{Bamdad \headeretal: SLAM for Visually Impaired People: a Survey}

\corresp{Corresponding author: Marziyeh Bamdad (e-mail: bamdad@ifi.uzh.ch).}

\begin{abstract}
In recent decades, several assistive technologies have been developed to improve the ability of blind and visually impaired (BVI) individuals to navigate independently and safely. 
At the same time, simultaneous localization and mapping (SLAM) techniques have become sufficiently robust and efficient to be adopted in developing these assistive technologies. 
We present the first systematic literature review of 54 recent studies on SLAM-based solutions for blind and visually impaired people, focusing on literature published from 2017 onward. This review explores various localization and mapping techniques employed in this context. 
We systematically identified and categorized diverse SLAM approaches and analyzed their localization and mapping techniques, sensor types, computing resources, and machine-learning methods.
We discuss the advantages and limitations of these techniques for blind and visually impaired navigation. Moreover, we examine the major challenges described across studies, including practical challenges and considerations that affect usability and adoption. Our analysis also evaluates the effectiveness of these SLAM-based solutions in real-world scenarios and user satisfaction, providing insights into their practical impact on BVI mobility.
The insights derived from this review identify critical gaps and opportunities for future research activities, particularly in addressing the challenges presented by dynamic and complex environments. We explain how SLAM technology offers the potential to improve the ability of visually impaired individuals to navigate effectively. Finally, we present future opportunities and challenges in this domain.
\end{abstract}


\titlepgskip=-21pt

\maketitle

%
%

\section{Introduction}
\label{sec:introduction}
\PARstart In recent decades, there has been increasing research interest in developing assistive technologies to enhance spatial navigation for blind and visually impaired (BVI) individuals. In most cases, the main goal is to guide and assist BVI people in navigating safely in unknown environments without the help of a sighted assistant. Navigation is a complex task; it requires finding an optimal path to the desired destination, perceiving the surroundings, and avoiding obstacles. Crucially, all of these functionalities need to accurately localize the BVI user in the environment. There are several approaches for localization, such as the global positioning system (GPS), radio frequency identification (RFID), and simultaneous localization and mapping (SLAM)~\cite{panigrahi2021localization, alkendi2021state}. Each has advantages and challenges and is used in different applications.

GPS is a localization technique employed in outdoor scenarios owing to its affordability to the end user, wide coverage of the Earth, and ease of integration with other technologies. However, this technique suffers from limitations like satellite signal blockage, inaccuracy, and signal loss caused by weather conditions, walls, and other obstacles~\cite{alkendi2021state}. 
Approaches based on RFID utilize small, low-cost tags for localization. To localize an agent, a set of RFID tags must be installed in the environment~\cite{panigrahi2021localization}. Although localization can be accurately performed using an RFID scheme, taking advantage of this technique requires a pre-installed infrastructure. 

A SLAM approach can offer a reliable alternative to RFID and GPS. SLAM is an innovative technique that involves simultaneously constructing an environment model (map) and estimating the state of an agent moving within it~\cite{cadena2016past}. The SLAM architecture (Figure \ref{slam_arch}) consists of two fundamental components: the front-end and the back-end. The front-end receives environmental information from the sensors, abstracts it into amenable models for estimation, and sends it to the back-end~\cite{cadena2016past}.
\Figure[t!](topskip=0pt, botskip=0pt, midskip=0pt){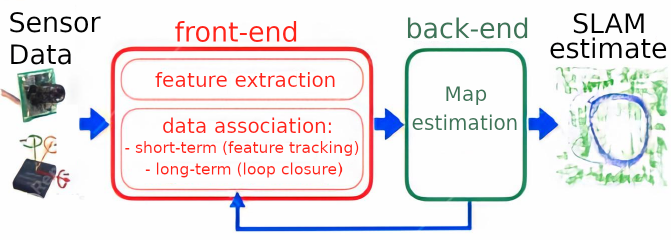}
{\textbf{Front- and back-end in a typical SLAM system~\cite{cadena2016past}}\label{slam_arch}}
The back-end is responsible for optimizing the mapping, localization, and data fusion processes, which collectively contribute to the accuracy and reliability of the SLAM systems. 

SLAM uses diverse types of sensors to determine an agent's position, location, and velocity and detect and avoid obstacles, even in a dynamically changing unknown environment. This technique uses infrared (IR) sensors, acoustic sensors, RGB cameras, inertial measurement units (IMUs), ultrawide-band (UWB), LiDAR, RADAR, and RGB-D sensors~\cite{khan2022investigation}.

The collaborative effort between the front- and back-ends empowers SLAM to provide a robust and real-time spatial understanding, making it a valuable tool for various applications.
The SLAM community has made tremendous strides over the past 35 years, developing large-scale practical applications and seeing a steady transition of this technology into the industry~\cite{cadena2016past}.

SLAM technology has been widely applied in various fields, demonstrating its versatility and robustness. In autonomous driving, SLAM enables vehicles to create and update maps of their environments in real time \cite{cheng2022review}. This capability is essential for safe and efficient road navigation, allowing accurate localization and obstacle avoidance in dynamic urban environments. 
In robotics, SLAM is crucial for navigation \cite{siegwart2011introduction}. This allows robots to operate effectively in unknown environments by simultaneously building a map of their surroundings and determining their positions within them. Such a capability is essential for autonomous exploration, path planning, and obstacle avoidance in diverse settings, ranging from indoor spaces to outdoor terrains. 
Augmented reality (AR) relies heavily on SLAM for the accurate placement of virtual objects in the real world, enhancing user experiences in gaming, education, and industrial applications \cite{jinyu2019survey}. 
In underwater robotics, SLAM helps autonomous underwater vehicles navigate highly unstructured and complex marine environments and supports exploration, research, and maintenance \cite{zhang2022visual}. 
For aerial vehicles, SLAM is an indispensable methodology for autonomous flight performed by unmanned aerial vehicles (UAVs) \cite{cioffi2023hdvio}, along with flight control \cite{xu2024adaptive, xu2023adaptive}. SLAM enables drones to navigate and map areas autonomously, which is valuable for tasks such as search-and-rescue, surveying, and environmental monitoring. These diverse applications highlight the versatility of SLAM and its critical role in enabling autonomous operation across various scenarios and platforms, from urban landscapes to ocean depths.

The evolution of portable computation and the availability of low-cost, highly accurate, and lightweight sensors such as cameras and IMUs make them appropriate for pedestrian navigation. By exploiting these advances, many researchers have recently adopted SLAM to develop assistive technology demonstrators to help BVI people navigate unknown environments. 

Since the first electronic travel aids (ETAs) emerged approximately 70 years ago, the development of navigation devices to guide BVI people through indoor and/or outdoor environments has remained a challenge and a key concern for researchers \cite{real2019navigation}.
From traditional to deep-learning-based navigation approaches, researchers have always faced challenges ranging from technical issues to the limitations of user capabilities. As BVI navigation approaches must improve real-time performance while reducing the size, weight, energy cost, and overall price of the assistive system, these studies have put a lot of effort into coping with constraints in computational issues, sensory equipment, and portable devices. They also need to provide solutions to calculate the precise position and orientation of the user in a real-time manner. 
However, the challenges of different scenarios, including complex and cluttered environments, noisy environments, and large spaces, must be considered.

Furthermore, efficient and reliable obstacle detection in both indoor and outdoor environments has always been a concern. In this regard, other challenges include identifying static and dynamic obstacles, predicting the risk of collision, understanding moving objects' motion and estimating their speed, detecting small objects, and identifying obstacles at different levels of the user's body, from drops in terrain to head level. 
In addition, an intuitive, user-friendly, low-cognitive-load method to provide accurate and sufficient environmental information to the user is also considered an important research target. These methods should be improved to provide adjustable and customized feedback on demand for different users. 

Moreover, assistive technology should provide user safety and independence, hands-free operations, decreased effort, and backup in the case of system failure. In addition to the aforementioned challenges, deep-learning-based solutions also have special issues, such as designing lightweight neural network architectures to reduce computational expense and provide sufficient data for the training and validation of the models.

This SLR is designed to act as a resource for the academic and research communities. The objective of this review is to explore and highlight the strengths and potential limitations of the current SLAM applications for visually impaired navigation. This study aims to inform and guide subsequent research. The insights derived from this review identify critical gaps and opportunities for future research, particularly for tackling the challenges presented by dynamic and complex environments. Such environments pose unique difficulties for visually impaired navigation, and addressing them through advanced SLAM technologies could lead to significant improvements in both the effectiveness and reliability of assistive solutions.

\subsection{Related Work}
Thus far, many reviews have been conducted on assistive technologies developed for BVI navigation. Several studies reviewed walking assistance systems~\cite{messaoudi2022review, parker2021wayfinding, khan2021analysis, romlay2021methodologies, zhang2021bibliometric, manjari2020survey, islam2019developing, real2019navigation, fernandes2019review} provided a detailed classification of the developed approaches. \cite{islam2019developing} categorized walking assistants into three groups: sensor-based, computer vision-based, and smartphone-based. The authors explained the technologies used and inspected each approach, and evaluated some important parameters of each approach, such as the type of capturing device, type of feedback, working area, cost, and weight.
The work by \cite{messaoudi2022review} introduced techniques and technologies designed to assist visually impaired individuals in their mobility and daily lives. This comprehensive review analyzes multiple mobility-assistive technologies that are suitable for indoor and outdoor environments. It offers insights into the various feedback methods employed by assistive tools based on recent technologies. 
In addition, \cite{parker2021wayfinding} reviewed wayfinding devices used by visually impaired individuals in real-world scenarios. This review aimed to provide a comprehensive exploration of the various aids employed for navigation while assessing their perceived efficacy.

Some studies focused on indoor navigation for BVI users~\cite{khan2022recent, wang2021survey, plikynas2020research, faccanha2020m, simoes2020review, plikynas2020indoor, kandalan2020techniques, zvironas2019indoor} and some focused on computer vision-based navigation systems~\cite{khan2022recent, walle2022survey, valipoor2022recent, fei2017review, sivan2016computer, jafri2014computer}. Among these studies,~\cite{khan2022recent} conducted a systematic literature review of state-of-the-art computer vision-based methods used for indoor navigation. The authors described the advantages and limitations of each solution under review, and included a brief description of each method.
Furthermore, \cite{kandalan2020techniques} comprehensively examined existing methods and systems developed within the domain of assistive technology, with a specific focus on addressing the unique needs and challenges faced by the visually impaired. This study places strong emphasis on evaluating methods that have practical applications in enhancing the lives of visually impaired individuals.

Several review papers on wearable navigation systems have also been published~\cite{xu2023wearable, hersh2022wearable, dos2021systematic, tapu2020wearable, chaudhary2019state, dakopoulos2009wearable}. \cite{xu2023wearable} have conducted a systematic review with the primary objectives of analyzing wearable obstacle avoidance electronic travel aids. Their work delves into the strengths and weaknesses of existing ETAs, providing a thorough evaluation of hardware functionality, cost-effectiveness, and the overall user experience. ~\cite{hersh2022wearable} provided a comprehensive understanding of wearable travel aids by focusing on their designs and usability. Their objectives included surveying the current landscape of travel aid design, investigating key design issues, and identifying limitations and future research directions. 
Furthermore,~\cite{dos2021systematic} conducted a systematic review of the literature on wearable technologies designed to enhance the orientation and mobility of the visually impaired. This review provides valuable insights into the technological characteristics of wearables, identifies feedback interfaces, emphasizes the importance of involving visually impaired individuals in prototype evaluations, and highlights the critical need for safety evaluations. 
A review by~\cite{tapu2020wearable} provides a comprehensive review of computer vision and machine-learning-based assistive methods. Existing ETAs are divided into two groups: active systems providing subject localization and object identification, and passive systems providing information about the users’ surroundings using a stereo camera, monocular camera, or RGB-D camera.

Focusing on guide robots,~\cite{thiyagarajan13intelligent} reviewed the multifaceted objectives. Their work included a comparative analysis of the existing robotic mobility aids and state-of-the-art technologies. This review highlights the potential of guide robots to enhance the mobility and independence of the visually impaired.

\cite{alam2018staircase} and \cite{kinra2023comprehensive} reviewed studies with the focus on  object detection and recognition. \cite{kinra2023comprehensive} performed a review on object recognition tailored to the needs of visually impaired individuals. This review examines state-of-the-art object detection and recognition techniques,  focuses on standard datasets, and emphasizes on the latest advancements. 
\cite{alam2018staircase} reviewed studies specific for staircase detection systems, primarily designed to facilitate the navigation of visually impaired individuals. The goal of this review is to provide a comprehensive comparative analysis of these systems considering their suitability and effectiveness.

Other similar studies include a survey of inertial measurement units (IMUs) in assistive technologies for visually impaired people~\cite{reyes2021inertial}, a review of urban navigation for BVI people~\cite{el2021systematic}, a survey paper that reviewed assistive tools based on white canes~\cite{motta2018overview}, and review papers exploring smartphone-based navigation devices~\cite{tan2022exploration, budrionis2022smartphone, khan2021insight}. ~\cite{tan2022exploration} reviewed the multifaceted objectives in the domain of smartphone-based navigation devices. They aimed to provide a comprehensive overview of smartphone use among people with vision impairment, identify research gaps for future exploration, and delve into the use of smartphones by individuals with vision impairment and the accessibility challenges they encounter.
To the best of our knowledge, there is no survey paper on SLAM-based navigation systems for BVI people. Our study aims to bridge this gap in literature.

\subsection{Contribution}
This paper presents a systematic literature review (SLR) addressing fundamental questions regarding SLAM-based approaches for BVI navigation. This review provides insights into technological diversity, advantages, limitations, and the potential to address real-world challenges. While recognizing the broad range of potential research questions we narrowed our focus to the four questions outlined in Table~\ref{research_questions}. The primary contributions of this study are as follows:
\begin{itemize}
    \item Identification of SLAM approaches: We systematically identified and categorized the diverse SLAM approaches adopted in the development of assistive systems tailored for visually impaired navigation. This includes analyzing the localization and mapping techniques, sensor types, computing resources, and machine-learning methods used in these approaches. 
    \item Advantages and limitations synthesis: Our study synthesizes the advantages and limitations of these SLAM techniques when applied to BVI navigation.  
    \item Classification of challenges: We identify and categorize studies that address challenging conditions relevant to SLAM-based navigation systems for the visually impaired. In addition, we discuss practical considerations that affect the usability and adoption of these systems. 
    \item Exploration of the potential for enhancing BVI navigation: We analyzed how the proposed SLAM-based approaches improved navigation in visually impaired individuals. In addition, we evaluated the effectiveness of these solutions in real-world scenarios and assessed user satisfaction to understand their practical impact on BVI mobility. 
\end{itemize}

\subsection{Paper structure}
The remainder of this paper is organized as follows. In Section \ref{sec:review_methodology}, we explain the protocol, methodology, tools, and techniques used to conduct SLR.
The findings of our SLR and answers to the SLR research questions are summarized in Section \ref{results}. 
Section \ref{future_opportunities} presents the future opportunities and potential advancements in this domain. Finally, Section \ref{conclusion} concludes the paper.
%
%
\section{SLR methodology}
\label{sec:review_methodology}
\begin{figure*}[t!]
    \centering
    \includegraphics[scale=0.5]{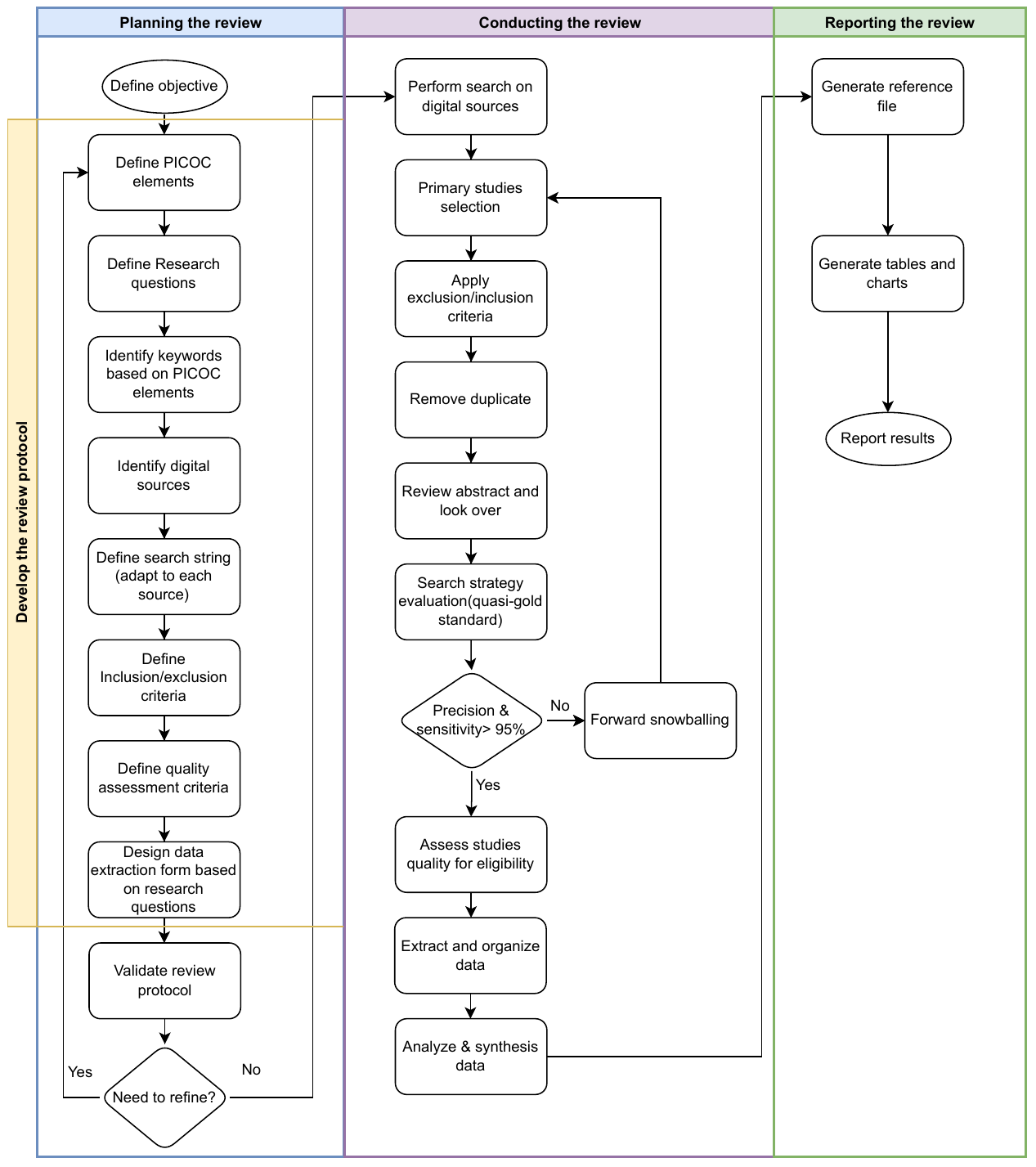}
    \caption{\textbf{The process of SLR}}
    \label{SLR_process}
\end{figure*}

A systematic literature review is one of the most common types of literature review used to collect, review, appraise, and report research studies on a specific topic, adhering to predefined rules for conducting the review \cite{grant2009typology}. Compared with traditional literature reviews, it provides a wider and more precise understanding of the topic under review \cite{pati2018write}. Various guidelines exist for conducting SLR in different research fields such as software engineering \cite{keele2007guidelines, kitchenham2009systematic, petersen2015guidelines}, computer science \cite{carrera2022conduct}, information systems \cite{okoli2010guide}, planning education and research \cite{xiao2019guidance}, and health sciences \cite {fink2019conducting, pati2018write}. To conduct this review, we followed the guidelines for conducting systematic reviews proposed by \cite{kitchenham2007guidelines}. Figure \ref{SLR_process} illustrates our SLR process.

SLR consists of three key phases: planning, conducting, and reporting the review.
We defined our research questions and motivation, keywords, and search string, as well as selection criteria in the planning phase of the SLR.
In the conducting phase, we executed searches on digital sources using predefined search strings that were established during the planning stage. We evaluated the quality of the selected papers and extracted relevant data aligned with SLR research questions.

We used the PICOC criteria to identify the key elements that needed to be considered and frame our research questions. PICOC represents Population, Intervention, Comparison, Outcome, and Context \cite{papaioannou2016systematic}. Table \ref{picoc} lists the PICOC elements, relevant values, and descriptions of these elements in our study.
%
%
\begin{table*}
    \centering
     \caption{\textbf{Elements of PICOC}}
     \label{picoc}
     \setlength{\tabcolsep}{3pt}
     \begin{tabularx}{\textwidth}{p{40pt} X X}
     \hline
     Element & Description \cite{papaioannou2016systematic} & Value \\
     \hline
    
    Population & The problem or situation  & Visually impaired navigation \\
    Intervention & The technology, tool or method under study  & SLAM \\
    Comparison (optional) & The technology, tool or method with which the intervention is compared & - \\
    Outcome & Results that Intervention could produce & Lightweight, affordable, accurate, efficient assistive technology \\
    Context & The specific context of the study & Autonomous mobility of visually impaired people \\
    \hline
    \end{tabularx}
\end{table*}

In this section, we first introduce the tool we used to manage our SLR process and then detail our methodology for conducting our systematic literature review in planning and conducting the review subsections.

%
%
\subsection{SLR Tool}
Various tools have been used to conduct systematic literature reviews. Some of them are commercial such as Covidence\footnote{https://www.covidence.org/home}, DistillerSR\footnote{https://www.evidencepartners.com/products/distillersr-systematic-review-software/}, and EPPI-Reviewer\footnote{https://eppi.ioe.ac.uk/CMS/Default.aspx}; and some are free such as Cadima\footnote{https://www.cadima.info/}, Rayyan\footnote{ https://rayyan.qcri.org/welcome}, RevMan\footnote{https://training.cochrane.org/online-learning/core-software-cochrane-reviews/revman}, and Parsifal\footnote{https://parsif.al}. We used the Parsifal platform to manage the SLR phases. It is an online tool developed to support the process of performing SLR. Parsifal provides researchers with an interface to invite co-authors to collaborate in a shared workspace on the SLR.
During the planning phase, this tool assists the authors by addressing the objectives, PICOC, research questions, search strings, keywords and synonyms, selection of sources, and inclusion and exclusion criteria. Parsifal offers tools for creating a quality assessment checklist and data extraction forms. In the conducting phase, this tool helps the authors import the bibtex files and select studies. It assists in identifying and eliminating duplicates among various sources, performing quality assessments, and facilitating data extraction from papers. Finally, it provides a method to document the entire SLR process.

%
%
\subsection{Planning the review}
The first step in conducting SLR is to establish a protocol. The protocol outlines the review procedures and ensures replicability. Within the protocol, we formulated our research questions, designed a search strategy, and defined the specific criteria for selecting relevant studies. In addition, we defined a set of criteria presented in Table \ref{quality_assessment} to evaluate the quality of the selected literature. Furthermore, to facilitate the extraction of data in alignment with our research questions, we designed a data-extraction form.

%
%
\subsubsection{Research Questions and Motivation}
The SLAM technique is widely used for the navigation of robots, autonomous drones, and self-driving cars, owing to its performance, reliability, and efficiency. Therefore, we reviewed the literature on visually impaired navigation, which was designed based on the SLAM technique. Our aim was to determine the advantages and limitations of employing this technique for visually impaired navigation as well as to identify opportunities for future research. Furthermore, we aimed to explore how extensively this method has been used in this specific area of research. Table~\ref{research_questions} presents the research questions that guided this review, and a description of  the questions.

%
%
\begin{table*}
    \centering
    \caption{\textbf{Research questions for the SLR process}}
     \label{research_questions}
     \setlength{\tabcolsep}{3pt}
     \begin{tabularx}{\textwidth}{p{15pt} X X}
    \hline
       No. & Research question & Description \\
       \hline
       RQ1 & What localization and mapping approaches are used for the navigation of the visually impaired? & The target is to identify different localization and mapping techniques adopted for the development of assistive devices for visually impaired navigation. \\
       
       RQ2 & What are the advantages and limitations of SLAM techniques for BVI navigation?  & The objective is to summarize the advantages and constraints of SLAM-based approaches for visually impaired navigation. \\
       
       RQ3 & What challenging situations have been addressed? & The purpose of this question is to know which of the challenging conditions (e.g. crowded environment, changing view point, challenging light conditions, dynamic objects, etc.) relevant to navigation systems have been considered. \\

        RQ4 & How does the proposed solution improve navigation using SLAM for individuals with impaired vision? & The research question seeks to understand how SLAM techniques can enhance mobility and navigation in individuals with visual impairments. \\
       
       \hline
    \end{tabularx}
\end{table*}

%
%
\subsubsection{Search strategy} \label{search_strategy}
A key step in performing SLR is to design an effective search strategy. This strategy should be executed with reasonable effort to retrieve relevant studies from digital libraries \cite{zhang2011identifying}. The exhaustive search process in systematic reviews is a critical factor distinguishing them from traditional literature reviews \cite{zhang2011identifying}, leading to a wider and more precise understanding of the topic under review.
To design the search string we first extracted keywords from the PICOC elements, including population, intervention, and outcomes. We then determined synonyms for each keyword to broaden the search string. The list of keywords and their synonyms related to each PICOC element is listed in Table \ref{keywords}.

%
%
\begin{table*}
    \centering
    \caption{\textbf{Keywords used to design search string}}
    \label{keywords}
    \setlength{\tabcolsep}{3pt}
    \begin{tabularx}{\textwidth}{p{90pt} X p{75px}}
         \hline
         Keywords & Synonyms & PICOC element \\
         \hline
         Visually impaired navigation & Blind navigation, Navigation assistance for the visually impaired, Navigational guidance for individuals with visual impairments, Navigational support for the visually disabled, Orientation and mobility for the visually impaired, Sight-impaired navigation, Visual impairment navigation aid, low vision navigation, partially sighted navigation & Population \\
         SLAM & Real-time mapping and positioning, Simultaneous Localization and Mapping, Simultaneous mapping and position tracking, mapping and localization & Intervention \\
         Accurate, efficient, reliable assistive technology & high-performance, precise, effective, trustworthy assistive technology & Outcome \\
         \hline
    \end{tabularx}
\end{table*}

The first part of the search string (i.e. 'visual* impair*' OR 'blind' OR 'visually disabled' OR 'sight impaired') is relevant to the population element of the PICOC framework. The addition of an asterisk to the terms 'visual' and 'impair' allows us to include various expressions, including 'visually impaired' and 'visual impairment.' 
The following segment of the query, consisting of 'navigation*' OR 'mobility' OR 'wayfinding,' also places emphasis on the population aspect within the context of the systematic review. The inclusion of an asterisk in 'navigation*' ensures comprehensive coverage, accounting for variations such as 'navigational.'
Regarding the Intervention component defined in the PICOC framework, we employed the term 'SLAM' in conjunction with synonyms identified in the literature from diverse domains where SLAM is applied, such as robotics, autonomous driving cars, and underwater SLAM. The last segment of the search string is connected to the outcome element of the PICOC.
Adding keywords such as "localization" alone or the specific names of SLAM techniques did not increase the number of related papers.
The search strings were employed on ten large citation databases, as listed in Table \ref{digital_libraries}, to carry out an exhaustive search. We modified the base search string according to the Search Tip in each library to satisfy specific requirements.

%
%

\begin{table*}
    \centering
    \caption{\textbf{Databases selected for the search procedure}}
    \label{digital_libraries}
    \setlength{\tabcolsep}{3pt}
    \begin{tabularx}{\textwidth}{X X X X}
         \hline
         Digital source & Web address & \# of papers & Last access date \\
         \hline
         
         ACM Digital Library & http://portal.acm.org & 4 & 22 Jul 2023 \\
         Google Scholar & https://scholar.google.com & 0 & 23 Jul 2023 \\
         IEEE Xplore & http://ieeexplore.ieee.org & 11 & 22 Jul 2023 \\
         MDPI & https://www.mdpi.com & 2 & 23 Jul 2023 \\
         PubMed & https://www.ncbi.nlm.nih.gov/pubmed & 3 & 22 Jul 2023 \\
         Science Direct & http://www.sciencedirect.com & 2 & 22 Jul 2023 \\
         Scopus & http://www.scopus.com & 22 & 23 Jul 2023 \\
         Springer Link & http://link.springer.com & 2 & 22 Jul 2023 \\
         Taley \& Francis & https://www.tandfonline.com & 0 & 22 Jul 2023 \\
         Wiley Online Library & https://onlinelibrary.wiley.com & 1 & 22 Jul 2023 \\
         \hline
    \end{tabularx}
\end{table*}

We utilized the Advanced Search feature in digital libraries to gain more control over our search parameters. The title, abstract, and keyword fields were selected to retrieve the search results. Searching on Google Scholar is somewhat different from searching for other digital libraries. Unlike other platforms, Google Scholar does not suggest various filters, requiring the manual incorporation of filters into the search string. Additionally, to identify English-written studies, we adjusted the language preference settings within our Google Scholar account to filter the search results in English.

%
%
\subsubsection{Selection criteria}

Table \ref{selection_criteria} presents the selection criteria used to identify the eligible studies during the selection process. 
The Availability criterion included studies accessible in full text from digital databases. In addition, the Language criterion ensured the inclusion of publications written only in English.
Furthermore, the Publication Period criterion restricted the inclusion of studies published between January 2017 and July 2023.
This timeframe was chosen to prioritize the most current and state-of-the-art approaches in this rapidly evolving field. By focusing on this recent period, we aimed to provide a comprehensive yet manageable review of the latest innovations without overwhelming readers with potentially outdated information.
The Type of Source criterion included conference and journal papers, which were considered peer-reviewed and academically recognized sources. Books, dissertations, newsletters, speeches, technical reports, and white papers were excluded. 
Finally, the Relevance criterion played an important role in the exclusion process; therefore, publications outside the scope of our study were excluded based on a review of their titles and abstracts.

%
%
\begin{table*}
    \centering
    \caption{\textbf{Selection criteria}}
    \label{selection_criteria}
    \setlength{\tabcolsep}{3pt}
    \begin{tabularx}{\textwidth}{X X X X}
         \hline
         Criteria & Inclusion & Exclusion \\
         \hline
         Availability & Available in full text & Not accessible in speciﬁc databases \\
         Language & English & Not written in English \\
         Publication period & From 2017 to July 2023 & Prior 2017 \\
         Type of source & Conferences and journals papers & Books, dissertation, newsletters, speeches, technical reports, white papers \\
         Relevance & Papers relevant to at least two research questions & Outside the scope of our research \\
         \hline
    \end{tabularx}
\end{table*}

%
%

\subsection{Conducting the review}
As shown in Figure \ref{SLR_process}, the review process began after the review protocol was finalized. The conducting phase is a multi-stage process that includes research identification, study selection, data extraction, and data synthesis.
In the research identification step, digital libraries were searched using adapted search strings that were specific to each library. This search aimed to collect a pool of potentially relevant primary studies.
The next step involved the selection of studies for which the relevance of each study to the review was evaluated. The steps involved in this process are illustrated in Figure. \ref{study_selection}.
During the data extraction phase, the data required from the studies were collected and analyzed.
We employed the data extraction form established during the development phase of the review protocol to ensure accurate extraction of information that addresses our research questions. 

\subsubsection{Identification}\label{identification}
During the initial phase of our review, we conducted searches across the digital libraries using custom-formulated queries for each library. For each dataset, we ran three different search strings (SS1, SS2, and SS3), as shown in Table \ref{tab:search_strings}, consisting of various combinations of keywords, booleans, and wildcard operators. These search strings were applied to all digital libraries except Google Scholar. For Google Scholar, we initially used keywords similar to those used in SS1, resulting in over 11,000 results. Upon reviewing a subset of these, we determined that a significant number were not relevant to our topic. Consequently, we decided to use only the primary keywords (shown in Table \ref{keywords}) to construct the search string for this digital library.

We selected the search string that yielded the most results to identify primary studies and then applied exclusion criteria to the results obtained from the search strings used for each library. We observed that SS3, which incorporated 'Orientation and mobility' to refine the search by focusing on more specialized literature, did not yield better results than SS1, which included the general term 'mobility', across all digital libraries. This indicates that the broader term 'mobility' was sufficient to capture the necessary literature. The specificity of SS3 did not contribute to additional relevant results. Additionally, upon receiving the message 'Use fewer Boolean connectors (maximum 8 per field)' while running SS1 on ScienceDirect, we switched to SS2 to maintain the number of Boolean connectors within the limit.

The initial searches of all digital libraries resulted in 6,809 records. The search strings used for each digital library is presented in Table \ref{tab:search_results}.

\begin{table*}[h]
  \centering
  \caption{\textbf{Search strings applied to digital libraries, featuring keywords and operators to identify primary studies.}}
  \label{tab:search_strings}
  \setlength{\tabcolsep}{3pt} 
  \renewcommand{\arraystretch}{1.5} 
  \begin{tabularx}{\textwidth}{| >{\centering\arraybackslash}X | >{\centering\arraybackslash}X | >{\centering\arraybackslash}X | >{\centering\arraybackslash}X |}
    \hline
    \textbf{SS1} & \textbf{SS2} & \textbf{SS3} & \textbf{SS4}\\
    \hline
    ("Visually impaired" OR blind OR "visual impairment*" OR "visually disabled" OR "Sight impaired") 
    AND (navigation* OR mobility) AND (SLAM OR "Simultaneous Localization and Mapping" OR "Real-time mapping and positioning" OR "Simultaneous mapping and position tracking" OR "mapping and localization") AND (technology OR aid OR support OR assist*)
    & 
    ("Visually impaired" OR blind) AND (navigation OR mobility) AND (SLAM OR "Simultaneous Localization and Mapping")
    & 
    ("Visually impaired" OR blind OR "visual impairment*" OR "visually disabled" OR "Sight Impaired") AND (navigation* OR \textbf{"Orientation and mobility"}) AND (SLAM OR "Simultaneous Localization and Mapping" OR "Real-time mapping and positioning" OR "Simultaneous mapping and position tracking" OR "mapping and localization") AND (technology OR aid OR support OR assist*)
    &
    ("conference paper" OR "journal") AND ("Visually impaired") AND  (navigation) AND (SLAM) AND (-review) AND ( -survey)
    \\
    \hline
  \end{tabularx}
\end{table*}
\begin{table}[h]
  \centering
  \caption{Utilization of search strings by digital libraries}
  \label{tab:search_results}
  \begin{tabular}{ l c c }
    \hline
    \textbf{Digital source} & \textbf{Search string} & \textbf{Number of results} \\
    \hline
    ACM Digital Library & SS1 & 284 \\
    Google Scholar & SS4 & 602 \\
    IEEE Xplore & SS2 & 50  \\
    MDPI & SS2 & 191 \\
    PubMed & SS1 & 8 \\
    Science Direct & SS2 & 518 \\
    Scopus & SS2 & 1510 \\
    Springer Link & SS2 & 2585  \\
    Taley \& Francis & SS2 & 486 \\
    Wiley Online Library & SS2 & 575 \\
    \hline
  \end{tabular}
\end{table}

The results obtained from digital libraries searches were exported in the BibTex format, a process facilitated by the export citation features available in the libraries. The BibTex data were then imported into the Parsifal framework for subsequent stages of our review.
Springer and Google Scholar do not provide direct options for exporting data in BibTex format. To address this issue, we used Zotero and its browser plugin, Zotero Connector, to streamline the process. With these tools, we added paper information from webpage views to Zotero and subsequently retrieved BibTex data.

For Springer Link, which provides only CSV files with search results, we opened the CSV in Excel and extracted the DOIs. These DOIs were then pasted into Zotero's "Add item(s) by identifier" feature. After importing the DOIs into Zotero, we selected the appropriate folder containing the imported papers and exported the collection to the BibTex format using a simple right-click.
As Scholar Google does not provide easy export of a large number of records, we adopted a similar approach: creating a library, saving search results to that library, and exporting paper data in BibTex format from that library. This process ensured that we obtained the necessary data for the subsequent stages of our systematic review.

\subsubsection{Study selection}
After conducting searches in the digital libraries, we applied our selection criteria, as defined in our review protocol, to filter out irrelevant studies. Initially, records published before 2017 were excluded. Further exclusions involved filtering out publications that were not written in English or had not been published in peer-reviewed venues. Following these steps, of the initial 6809 records found in the initial search, 5431 were excluded.

We imported the study data into the Parsifal platform in BibTeX format, as explained in Section \ref{identification}, which helped remove duplicate studies. A total of 116 duplicate papers were excluded. We then reviewed the titles and abstracts of the remaining studies, excluding those irrelevant to our research topic. In this step, 779 studies were excluded.

In the next step, we performed a fast reading of the full text of the remaining papers, excluding 265 studies that were outside the scope of our research. We then evaluated the quality of the studies based on the quality assessment criteria defined in the SLR protocol. Five studies were removed during the assessment of study quality. Table \ref{quality_assessment} lists the quality assessment criteria for our SLR.

%
%
\begin{table}
     \caption{\textbf{Quality assessment criteria and weights}}
     \label{quality_assessment}
     \setlength{\tabcolsep}{3pt}
     \begin{tabular}{p{170pt} p{45pt}}
    \hline
       Criteria & Weight \\
       \hline
       Is there an adequate description of the context in which the research was carried out? & {0.0, 0.5, 1.0} \\
       Does the methodology take into account both localization and mapping issues? & {0.0, 0.1, 2.0} \\
       Is the solution proposed well presented? & {0.0, 0.5, 1.0} \\
       Is there a clear statement of findings? & {0.0, 0.5, 1.0} \\
       Is the research design appropriate to address the aims of the research? & {0.0, 0.5, 1.0} \\
       Does the study add value to the research community? & {0.0, 0.5, 1.0} \\
       \hline
    \end{tabular}
\end{table}

We carefully read 213 full-text articles to address the research questions. As 166 articles were not relevant to at least two of our research questions, they were removed, leaving 47 articles for the final stage.

To objectively assess the performance of our search strategy, we employed the quasi-gold standard (QGS) technique, as described by \cite{zhang2011identifying}. Using this method, a set of articles related to the research topic is manually selected. Digital libraries are then searched based on the research strategy to identify related studies. Finally, the retrieved articles are compared with QGS, and the sensitivity of the search strategy is calculated using the following formula:
\[
\text{Sensitivity} = \dfrac{\text{Number of relevant studies retrieved}}{\text{Total number of relevant studies}} \times 100
\]
In our SLR, with 30 manually selected relevant studies and 48 studies retrieved using the SLR search strategy, of which 26 were among the manually selected studies, the resulting sensitivity was approximately 86.67\%.

\begin{figure}
    \centering
    \includegraphics[scale=0.5]{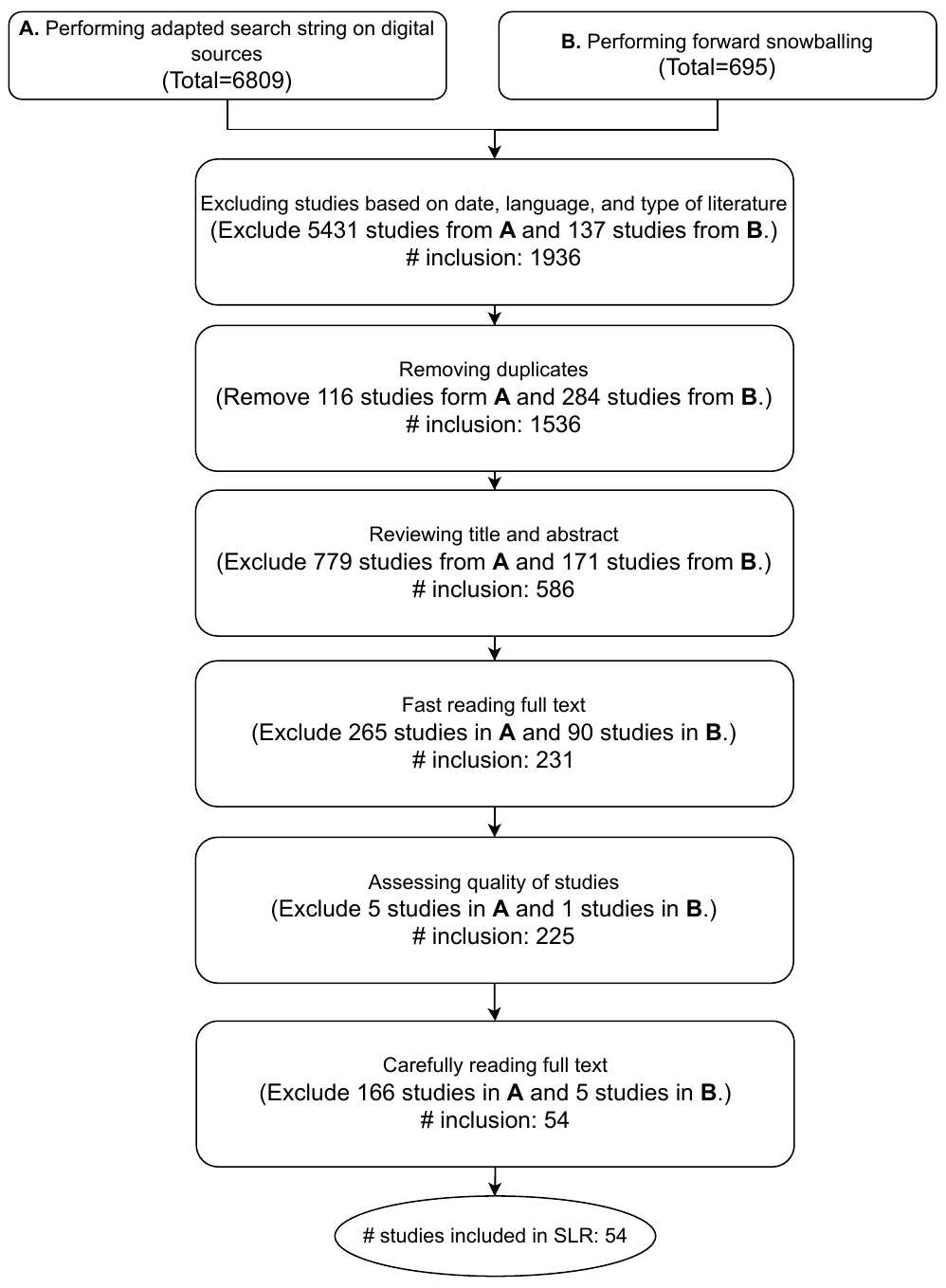}
    \caption{\textbf{Studies selection process}}
    \label{study_selection}
\end{figure}

To provide a broader range of relevant studies, we included papers on the forward snowballing process \cite{wohlin2014guidelines}. This process involves identifying and accessing references in a paper and reviewing cited papers. We used "Cited by" feature of Google Scholar to identify these additional papers.
In this stage, 695 articles were identified. After removing duplicates and applying selection criteria similar to those used for the articles obtained from digital libraries, we added seven more articles to the final collection. Consequently, 54 articles were included in this review. Figure \ref{study_selection} shows a diagram of the study-selection process.

It is important to note that the last search conducted in digital libraries was on July 23, 2023, and that for forward snowballing was on August 12, 2023. These dates should be considered as the starting points for future reviews.
The publications included in our review are listed in Tables \ref{publications_part1}–\ref{publications_part3} and categorized based on their publication venues. These tables provide an overview of the literature, including the paper title, author names, publication year, location, and source through which the studies were discovered. Among the 54 studies included in our SLR, 27 were sourced from journals, as presented in Table \ref{publications_part1}. The remaining 27 studies were presented at conferences, as shown in Tables \ref{publications_part2} and \ref{publications_part3}.

Additionally, Tables \ref{tab:t_limitations_potential_solutions_partI}–\ref{tab:t_limitations_potential_solutions_partIV} summarize the perspectives and innovations presented in the publications with insights into their limitations and advantages. These tables demonstrate the research issues addressed and the contributions of each study, highlighting the strengths and potential drawbacks of the proposed solutions. They also indicate which solutions are open source, with only seven papers having some or all parts of the project available as open source. Links to the sources are provided in these tables if they are directly available in the relevant papers.

\subsubsection{Data extraction}
Data extraction is a critical phase in the systematic literature review process in which relevant data from selected studies are systematically collected. To achieve this objective, we employed the data-extraction form defined in the SLR protocol. This form consists of various fields designed to retrieve answers to our research questions from each of the included articles. Within the scope of this SLR, we defined the following essential elements, each contributing to a comprehensive understanding of the reviewed literature:

\begin{itemize}
    \item Short summary of the paper: A concise overview of the main points and findings of the study.
    \item Research issue and contribution: Summary of the research issues addressed and contributions of the studies.
    \item Localization and mapping technique: Identification of specific techniques applied for localization and mapping.
    \item Localization and mapping accuracy and robustness: Assessment of accuracy and robustness levels in localization and mapping techniques.
    \item Running time: Analysis of the running time for localization and mapping techniques.
    \item Advantages of the presented method: The strengths associated with the localization method presented in each paper for visually impaired navigation.
    \item Limitations of the presented method: Identification of weaknesses or constraints associated with the localization technique.
    \item Types of obstacles addressed: Categorization of obstacles, static and dynamic, as a challenge during navigation.
    \item Challenging conditions: Explanation of other challenging scenarios that the methods are designed to handle.
    \item Types of sensors: Identification of sensors employed to receive data from the surroundings.
    \item Computing resources: Identification of computing resources used in SLAM-based solutions.
    \item Improvement in navigation: Identification of how SLAM-based methods enhance navigation for individuals with impaired vision.
    \item Working area: Whether the method is intended for indoor, outdoor, or both indoor and outdoor environments.
    \item Practical challenges and operational efficiency: Evaluation of the user-friendliness, cost-efficiency, weight, comfort for extended use, adjustable fit, fatigue mitigation, and portability of the SLAM-based assistive tools.
    \item System prototype information: Detailed information on functionalities, sensors, computing resources, human-computer interaction (HCI) mechanisms, assistive tools, and battery life.
    \item User evaluation: Assessment of user satisfaction of the SLAM-based assistive tools.
    \item Machine learning techniques: Identification of machine learning techniques used in assistive solutions.
    \item Open-source availability: Identification of open-source contributions in the reviewed studies.
    \item Possible future opportunities and directions: Exploration of potential future research areas and directions stemming from these findings.
    \item The research questions addressed: Identification of specific SLR research questions addressed by each study.
\end{itemize}
\begin{table*}
 \scriptsize
\centering
\caption{\textbf{List of journal publications included in SLR}}
\label{publications_part1}
\setlength{\tabcolsep}{3pt}
\begin{tabularx}{\textwidth}{p{20pt} X X p{60pt} p{15pt} p{40pt}}
\hline
\textbf{Ref.} & \textbf{Title} & \textbf{Authors} & \textbf{Published} & \textbf{Year} & \textbf{Source} \\
\hline
\cite{ahmetovic2023sonification} & Sonification of navigation instructions for people with visual impairment & Dragan Ahmetovic and Federico Avanzini and Adriano Baratè and Cristian Bernareggi and Marco Ciardullo and Gabriele Galimberti and Luca A. Ludovico and Sergio Mascetti and Giorgio Presti & International Journal of Human-Computer Studies & 2023 & Science@Direct \\
\cite{li2022sensing} & Sensing and Navigation of Wearable Assistance Cognitive Systems for the Visually Impaired & Li, Guoxin and Xu, Jiaqi and Li, Zhijun and Chen, Chao and Kan, Zhen & IEEE Transactions on Cognitive and Developmental Systems & 2023 & IEEE Digital Library \\
\cite{song2023mixture} & Mixture reality-based assistive system for visually impaired people & Jucheng Song and Jixu Wang and Shuliang Zhu and Haidong Hu and Mingliang Zhai and Jiucheng Xie and Hao Gao & Displays & 2023 & Science@Direct \\
\cite{zhang2023research} & Research on Design and Motion Control of a Considerate Guide Mobile Robot for Visually Impaired People & Zhang, Bin and Okutsu, Mikiya and Ochiai, Rin and Tayama, Megumi and Lim, Hun-Ok & IEEE Access & 2023 & Scopus \\
\cite{yang2022unav} & UNav: An Infrastructure-Independent Vision-Based Navigation System for People with Blindness and Low Vision & Yang, Anbang and Beheshti, Mahya and Hudson, Todd E and Vedanthan, Rajesh and Riewpaiboon, Wachara and Mongkolwat, Pattanasak and Feng, Chen and Rizzo, John-Ross & Sensors & 2022 & PubMed \\
\cite{zhou2022multi} & Multi-Floor Indoor Localization Based on Multi-Modal Sensors & Zhou, Guangbing and Xu, Shugong and Zhang, Shunqing and Wang, Yu and Xiang, Chenlu & Sensors & 2022 & Scopus \\
\cite{hou2022knowledge} & Knowledge driven indoor object-goal navigation aid for visually impaired people & Hou, Xuan and Zhao, Huailin and Wang, Chenxu and Liu, Huaping & Cognitive Computation and Systems & 2022  & Wiley Online Library \\
\cite{plikynas2022indoor} & Indoor-Guided Navigation for People Who Are Blind: Crowdsourcing for Route Mapping and Assistance & Plikynas, Darius and Indriulionis, Audrius and Laukaitis, Algirdas and Sakalauskas, Leonidas & Applied Sciences (Switzerland) & 2022 & Scopus \\
\cite{xie2022multi} & A Multi-Sensory Guidance System for the Visually Impaired Using YOLO and ORB-SLAM & Xie, Zaipeng and Li, Zhaobin and Zhang, Yida and Zhang, Jianan and Liu, Fangming and Chen, Wei & Information & 2022 & MDPI \\
\cite{qiu2022egocentric} & Egocentric Human Trajectory Forecasting with a Wearable Camera and Multi-Modal Fusion & Qiu, Jianing and Chen, Lipeng and Gu, Xiao and Lo, Frank P.-W. and Tsai, Ya-Yen and Sun, Jiankai and Liu, Jiaqi and Lo, Benny & IEEE Robotics and Automation Letters & 2022 & Scopus \\
\cite{chen2021wearable} & A wearable navigation device for visually impaired people based on the real-time semantic visual slam system & Chen, Zhuo and Liu, Xiaoming and Kojima, Masaru and Huang, Qiang and Arai, Tatsuo & Sensors & 2021 & PubMed \\
\cite{slade2021multimodal} & Multimodal sensing and intuitive steering assistance improve navigation and mobility for people with impaired vision & Slade, Patrick and Tambe, Arjun and Kochenderfer, Mykel J. & Science Robotics & 2021 & Scopus \\
\cite{lu2021assistive} & Assistive Navigation Using Deep Reinforcement Learning Guiding Robot With UWB/Voice Beacons and Semantic Feedbacks for Blind and Visually Impaired People & Lu, Chen-Lung and Liu, Zi-Yan and Huang, Jui-Te and Huang, Ching-I and Wang, Bo-Hui and Chen, Yi and Wu, Nien-Hsin and Wang, Hsueh-Cheng and Giarré, Laura and Kuo, Pei-Yi & Frontiers in Robotics and AI & 2021 & Scopus \\
\cite{hakim2021indoor} & Indoor Wearable Navigation System Using 2D SLAM Based on RGB-D Camera for Visually Impaired People & Hakim, Heba and Fadhil, Ali & Advances in Intelligent Systems and Computing & 2021 & Scopus \\
\cite{zhang2021rgb} & An RGB-D Camera Based Visual Positioning System for Assistive Navigation by a Robotic Navigation Aid & Zhang, He and Jin, Lingqiu and Ye, Cang & IEEE/CAA Journal of Automatica Sinica & 2021 & IEEE Digital Library \\
\cite{cheng2021hierarchical} & Hierarchical visual localization for visually impaired people using multimodal images & Cheng, Ruiqi and Hu, Weijian and Chen, Hao and Fang, Yicheng and Wang, Kaiwei and Xu, Zhijie and Bai, Jian & Expert Systems with Applications & 2021 & Scopus \\
\cite{liu2020indoor} & Indoor Topological Localization Based on a Novel Deep Learning Technique & Liu, Qiang and Li, Ruihao and Hu, Huosheng and Gu, Dongbing & Cognitive Computation & 2020 & Scopus \\
\cite{jin2020combining} & Combining Obstacle Avoidance and Visual Simultaneous Localization and Mapping for Indoor Navigation & Jin, SongGuo and Ahmed, Minhaz Uddin and Kim, Jin Woo and Kim, Yeong Hyeon and Rhee, Phill Kyu & Symmetry & 2020 & MDPI \\
\cite{bai2019wearable} & Wearable travel aid for environment perception and navigation of visually impaired people & Bai, Jinqiang and Liu, Zhaoxiang and Lin, Yimin and Li, Ye and Lian, Shiguo and Liu, Dijun & Electronics (Switzerland) & 2019 & Scopus \\
\cite{zhang2019arcore} & An ARCore based user centric assistive navigation system for visually impaired people & Zhang, Xiaochen and Yao, Xiaoyu and Zhu, Yi and Hu, Fei & Applied Sciences (Switzerland) & 2019 & Scopus \\
\cite{bai2018virtual} & Virtual-Blind-Road Following-Based Wearable Navigation Device for Blind People & Bai, Jinqiang and Lian, Shiguo and Liu, Zhaoxiang and Wang, Kai and Liu, Dijun & IEEE Transactions on Consumer Electronics & 2018 & IEEE Digital Library \\
\cite{zhang2017indoor} & An indoor wayfinding system based on geometric features aided graph SLAM for the visually impaired & Zhang, He and Ye, Cang & IEEE Transactions on Neural Systems and Rehabilitation Engineering & 2017 & PubMed \\
\cite{zhang2020plane} & Plane-Aided Visual-Inertial Odometry for 6-DOF Pose Estimation of a Robotic Navigation Aid & Zhang, He and Ye, Cang & IEEE Access & 2020 & Scopus \\
\cite{albogamy2021sravip} & SRAVIP: Smart Robot Assistant for Visually Impaired Persons & Albogamy, Fahad and Alotaibi, Turk and Alhawdan, Ghalib and Mohammed, Faisal & International Journal of Advanced Computer Science and Applications & 2021 & Forward Snowballing \\
\cite{crabb2023lightweight} & A Lightweight Approach to Localization for Blind and Visually Impaired Travelers & Crabb, Ryan and Cheraghi, Seyed Ali and Coughlan, James M & Sensors & 2023 & Forward Snowballing \\
\cite{son2022wearable} & Wearable system to guide crosswalk navigation for people with visual impairment & Son, Hojun and Weiland, James & Frontiers in Electronics & 2022 & Forward Snowballing \\
\cite{hakim2019indoor} & Indoor Low Cost Assistive Device using 2D SLAM Based on LiDAR for Visually Impaired People & Hakim, Heba and Fadhil, Ali & Iraqi Journal for Electrical \& Electronic Engineering & 2019 & Forward Snowballing \\
\hline
\end{tabularx}
\end{table*}
\begin{table*}
\footnotesize
\centering
\caption{\textbf{List of conference papers included in SLR - part 1.}}
\label{publications_part2}
\setlength{\tabcolsep}{3pt}
\begin{tabularx}{\textwidth}{p{20pt} X X p{60pt} p{15pt} p{40pt}}
\hline
\textbf{Ref.} & \textbf{Title} & \textbf{Authors} & \textbf{Published} & \textbf{Year} & \textbf{Source} \\
\hline
\cite{goswami2023efficient} & Efficient Real-Time Localization in Prior Indoor Maps Using Semantic SLAM & Goswami, R. G. and Amith, P. V. and Hari, J. and Dhaygude, A. and Krishnamurthy, P. and Rizzo, J. and Tzes, A. and Khorrami, F. & 9th Inter. Conf. on Automation, Robotics and Applications (ICARA) & 2023 & IEEE Digital Library \\
\cite{hao2022detect} & Detect and Approach: Close-Range Navigation Support for People with Blindness and Low Vision & Hao, Yu and Feng, Junchi and Rizzo, John-Ross and Wang, Yao and Fang, Yi & European Conf. on Computer Vision & 2022 & Springer Link \\
\cite{kuribayashi2023pathfinder} & PathFinder: Designing a Map-Less Navigation System for Blind People in Unfamiliar Buildings & Kuribayashi, Masaki and Ishihara, Tatsuya and Sato, Daisuke and Vongkulbhisal, Jayakorn and Ram, Karnik and Kayukawa, Seita and Takagi, Hironobu and Morishima, Shigeo and Asakawa, Chieko & CHI Conf. on Human Factors in Computing Systems & 2023 & ACM Digital Library \\
\cite{agrawal2022novel} & A Novel Perceptive Robotic Cane with Haptic Navigation for Enabling Vision-Independent Participation in the Social Dynamics of Seat Choice & Agrawal, Shivendra and West, Mary Etta and Hayes, Bradley & IEEE Inter. Conf. on Intelligent Robots and Systems & 2022 & Scopus \\
\cite{rui2021multi} & A Multi-Sensory Blind Guidance System Based on YOLO and ORB-SLAM & Rui, Chufan and Liu, Yichen and Shen, Junru and Li, Zhaobin and Xie, Zaipeng & IEEE Inter. Conf. on Progress in Informatics and Computing (PIC) & 2021 & IEEE Digital Library \\
\hline
\end{tabularx}
\end{table*}
\begin{table*}[t]
 \scriptsize
\centering
\caption{\textbf{List of conference papers included in SLR - part 2.}}
\label{publications_part3}
\setlength{\tabcolsep}{3pt}
\begin{tabularx}{\textwidth}{p{20pt} X X p{60pt} p{15pt} p{40pt}}
\hline
\textbf{Ref.} & \textbf{Title} & \textbf{Authors} & \textbf{Published} & \textbf{Year} & \textbf{Source} \\
\hline
\cite{eden2018indoor} & Indoor Navigation using Text Extraction & Eden, Jake and Kawchak, Thomas and Narayanan, Vijaykrishnan & IEEE Inter. Workshop on Signal Processing Systems (SiPS) & 2018 & IEEE Digital Library \\
\cite{mulky2018autonomous} & Autonomous Scooter Navigation for People with Mobility Challenges & Mulky, Rajath Swaroop and Koganti, Supradeep and Shahi, Sneha and Liu, Kaikai & IEEE Inter. Conf. on Cognitive Computing (ICCC) & 2018 & IEEE Digital Library \\
\cite{lalonde2018localizing} & Localizing people in crosswalks using visual odometry: Preliminary results & Lalonde, Marc and St-Charles, Pierre-Luc and Loupias, Délia and Chapdelaine, Claude and Foucher, Samuel & Inter. Conf. on Pattern Recognition Applications and Methods (ICPRAM) & 2018 & Scopus \\
\cite{zhang2017plane} & Plane-aided visual-inertial odometry for pose estimation of a 3D camera based indoor blind navigation & Zhang, He and Ye, Cang & British Machine Vision Conf. (BMVC) & 2017 & Scopus \\
\cite{chen2017ccny} & CCNY Smart Cane & Chen, Qingtian and Khan, Muhammad and Tsangouri, Christina and Yang, Christopher and Li, Bing and Xiao, Jizhong and Zhu, Zhigang & IEEE 7th Annual Inter. Conf. on CYBER Technology in Automation, Control, and Intelligent Systems (CYBER) & 2017 & IEEE Digital Library \\
\cite{bai2017cloud} & A Cloud and Vision-Based Navigation System Used for Blind People & Bai, Jinqiang and Liu, Dijun and Su, Guobin and Fu, Zhongliang & Inter. Conf. on artificial intelligence, automation and control technologies & 2017 & ACM Digital Library \\
\hline
\end{tabularx}
\end{table*}

\begin{table*}[t]
 \scriptsize
\centering
\caption{\textbf{A summary of perspectives and innovations in SLAM-based navigation solutions, with insights into limitations and advantages - Part I.}}
\label{tab:t_limitations_potential_solutions_partI}
\setlength{\tabcolsep}{3pt}
\begin{threeparttable}
\begin{tabularx}{\textwidth}{p{20pt} >{\centering\arraybackslash}p{40pt} X X X X X }
\hline
\textbf{Ref.} & \textbf{Open-source} & \textbf{Research Issue} & \textbf{Contribution} & \textbf{Advantages}  & \textbf{Limitations} \\
\hline

\cite{ahmetovic2023sonification} & $\bullet$ & Addressing sonification techniques for navigation instruction & Innovative sonification for BVI navigation & Publicly available localization approach & Prone to drift, utilizing markers for user positioning  \\
\cite{li2022sensing} & $\bullet$ & Enhancing indoor navigation for BVI with wearable technologies & Low-cost wearable, SLAM navigation, multitarget recognition & Non-intrusive wearable device, Map reuse, & Network impact, cognitive feedback needs \\
\cite{song2023mixture} & $\bullet$ & Improving perception and independent task completion for BVI & Mixed reality, real-time perception, remote assistance & Improves perception, functions in diverse indoors, advanced interactive platform, sensor integration & Network status influence \\
\cite{zhang2023research} & $\bullet$ & Addressing guide robots' lack of consideration for user status and obstacle properties & Introduced considerate robot design and spatial risk map for navigation & Robot adapts speed and movement, enhancing natural interaction without disturbing others; calculates pedestrian directions & Needs improved speech recognition, real-world testing with BVI, effective environmental interaction \\
\cite{yang2022unav} & $\bullet$ & Addressing limitations of sensor-based navigation for BVI & Infrastructure-independent vision-based navigation for BVI & Map-evolution feedback loop ensures dynamic updates, and offline computation allows for continued use during signal loss. & Requires a dense reference image database; difficult to adapt to diverse environments. \\
\cite{zhou2022multi} & $\bullet$ & Improving high-precision indoor localization in complex multi-floor environments & Hybrid localization framework combining visual and wireless signals for high-precision indoor localization & Multi-modal sensor integration, fusion-based localization, and GAN-based approach for efficient, high-precision multi-floor localization & Implementation complexity, low positioning accuracy, and dependence on fingerprint database and offline maps \\
\cite{hou2022knowledge} & $\bullet$ & Applying object-goal navigation to aid BVI in indoor settings & Migration of object-goal navigation to assistive devices and a knowledge-driven approach & Active assistance with context-aware, knowledge-driven navigation for improved indoor object-goal guidance & Dependency on unlabelled images, scene understanding complexity, and generalization to unfamiliar environments. \\
\cite{plikynas2022indoor} & $\bullet$ & Operationalizing Web 2.0 social networking for indoor navigation assistance & Innovative integration of Crowdsourcing and social networking for indoor navigation & Constant access to a 24/7 indoor route database, offline functionality, and flexible, user-friendly wearable devices & Relies heavily on social networking, high processing power, and stable Internet connectivity \\
\cite{xie2022multi} & $\bullet$ & Enhancing guidance systems for BVI with multi-sensory integration for improved indoor navigation & Integration of ORB-SLAM with YOLO, dense map generation, and practical prototype implementation & Obstacle avoidance with multi-sensory feedback, dense navigation maps, and real-time target detection, implemetation of a practical smart cane & Dense map may not always align with reality, pathfinding has high computational costs, and target detection is trained based on a generic dataset.
 \\
\cite{qiu2022egocentric} & \href{https://github.com/Jianing-Qiu/TISS}{Trajectory dataset} \tnote{1} &  Forecasting egocentric camera wearers' trajectories in crowded spaces & A new egocentric dataset, a Transformer-based model with cascaded cross-attention, and demonstration of socially compliant robot navigation & An egocentric view with multi-modal fusion for trajectory forecasting; Socially compliant robot navigation and assists visually impaired individuals. & Not mentioned \\
\cite{chen2021wearable} & $\bullet$ &  limited navigation options, wearable devices, semantic visual SLAM, cost-effective solutions & Semantic SLAM integration, real-time solution, efficient resource allocation & Real-time semantic understanding, efficient resource allocation, enhanced navigation accuracy, voice broadcast for destination assistance & Remote server challenges including internet access dependency, security, and privacy concerns \\
\cite{slade2021multimodal} & $\checkmark$ & Decreasing cognitive burden, increasing walking speed with a multimodal augmented cane & Improving mobility with sensors, intuitive steering, and advanced navigation features, increasing walking speed & Enhanced mobility, faster walking, precise steering, reduced cognitive load, fewer collisions, increased confidence, reliable obstacle avoidance, failure backup,  selecting preferred walking speed by user & Heavy, requiring mechanical assembly \\
\cite{lu2021assistive} & $\bullet$ & Enhancing navigation with improved robotic assistance in dynamic environments. & A novel haptic-guided robot with enhanced navigation, dynamic obstacle avoidance via UWB, and voice-enabled beacon feedback & Enhanced environmental information, intuitive navigation, precise UWB positioning, semantic feedback, and DRL-based obstacle avoidance. & Dynamic obstacles impact SLAM accuracy; implicit learning in the simulator may cause real-world uncertainties. \\
\cite{hakim2021indoor} & $\bullet$ & Efficient indoor navigation aid. & Multi sensor utilization, efficient algorithm utilization, and real-time voice guidance. & Cost-effective, real-time navigation, accurate mapping, obstacle detection, efficient path planning. & Limited to static, simple environments. \\
\cite{zhang2021rgb} & $\bullet$ & Enhancing navigation accuracy and robustness in large indoor spaces for assistive technologies. & Introduced DVIO for enhanced 6-DOF pose estimation and VPS for accurate assistive navigation. & Enhanced pose accuracy, real-time updates, effective wayfinding, obstacle avoidance, and significant pose error reduction. & Pose drift remains an issue in complex environments. \\

\hline
\end{tabularx} 
     \begin{tablenotes}
    \item[1] \url{https://github.com/Jianing-Qiu/TISS} (Accessed on 3 May 2024)
    \end{tablenotes}
\end{threeparttable}
\end{table*}
\begin{table*}[t]
 \scriptsize
\centering
\caption{\textbf{A summary of perspectives and innovations in SLAM-based navigation solutions, with insights into limitations and advantages - Part II.}}
\label{tab:t_limitations_potential_solutions_partII}
\setlength{\tabcolsep}{3pt}
\begin{threeparttable}
\begin{tabularx}{\textwidth}{p{20pt} >{\centering\arraybackslash}p{40pt} X X X X X }
\hline
\textbf{Ref.} & \textbf{Open-source} & \textbf{Research Issue} & \textbf{Contribution} & \textbf{Advantages}  & \textbf{Limitations} \\
\hline
\cite{cheng2021hierarchical} & $\bullet$ & Improving visual localization in challenging outdoor settings. & Unified Dual Desc network enhancing descriptor extraction and multimodal integration for assistive localization. & Enhances robustness using multimodal imaging, advanced descriptors, and sequential integration for outdoor navigation. & Lacks real-time execution capability in assistive device. \\
\cite{liu2020indoor} & $\bullet$ & Enhancing independence for visually impaired through semantic environmental understanding and navigation. & Integrates ConvNets for semantic mapping; enhances topological localization. & Enhanced accuracy, semantic guidance, robustness to changes. & Computational inefficiency, suboptimal performance in motion blur effects. \\
\cite{jin2020combining} & $\bullet$ & Addressing the lack of comprehensive solutions for indoor navigation and obstacle avoidance. & Integration of dynamic person-detection method (EER–ASSL) and VSLAM for real-time navigation assistance in cluttered environments. & Enhanced smooth movement, reliable obstacle avoidance, effective navigation in dynamic environments. & Limited instruction capabilities, the decrease in person detection performance under varying lighting conditions and speed. \\
\cite{bai2019wearable} & $\bullet$ & Addressing lack of integrated navigation and object recognition systems. & Integration of lightweight CNN-based object recognition and visual SLAM for improved environment perception and navigation. & Load cognitive decrease, safe and quick navigation, enhanced perception, and real-time performance on smartphones. & Unable to detect small-size obstacles and obstacle detection limitations. \\
\cite{zhang2019arcore} & $\bullet$ & Addressing lack of user-centric indoor navigation aids for visually impaired. & ARCore integration, adaptive path planning, and dual-channel user interaction for indoor navigation. & Enhanced mapping, obstacle-avoiding path planning, and intuitive dual-channel interaction for improved indoor navigation. & Reliance on existing indoor scenario CAD maps. \\
\cite{bai2018virtual} & $\bullet$ & Addressing gaps in localization, way-finding, and route following for visually impaired navigation. & Dynamic subgoal route following, visual SLAM integration, and wearable optical see-through glasses for enhanced indoor navigation. & Enhances precision with visual SLAM, cost-effective sensors, efficient obstacle avoidance, safe indoor navigation with dynamic subgoal selection. & Not mentioned \\
\cite{zhang2017indoor} & $\bullet$ & Addressing accumulative pose error, GPS-denied navigation, and real-time pose estimation. & New 6-DOF pose estimation method using floor and wall information, and a real-time wayfinding system. & Reduces accumulative pose error and provides real-time wayfinding  & System less effective in simple tasks, weight causes discomfort, and fails at high walking speeds over 0.6 m/s. \\
\cite{zhang2020plane} & $\bullet$ & Addressing accurate 6-DOF pose estimation challenges through innovative visual-inertial odometry for robotic navigation aids. & A plane-aided VIO method and a plane-consistency check for enhanced pose estimation accuracy. & Improved accuracy, a plane-consistency check, practical implementation for assistive navigation, and outperformance of state-of-the-art methods. & Not mentioned \\
\cite{albogamy2021sravip} & $\bullet$ & Enhancing navigation in an indoor public environment for pre-scheduled tasks with user-independent robotic solutions. & Innovative user-independent robotic assistance for indoor navigation. & Enhancing inclusivity and efficiency: user-independent robotic assistance. & It is not mentioned how the proposed approach handles crowds in the public places under study. \\
\cite{crabb2023lightweight} & \href{https://www.openicpsr.org/openicpsr/project/183714/version/V1/view}{ICOSR repository} \tnote{2} & Developing a lightweight indoor localization system using a 2D floor plan of the environment rather than a 3D model. & Innovative localization algorithm integrating visual landmarks, VIO, and 2D floor plans. & Smartphone-based, lightweight, and robust localization approach. & Noisy distance estimates due to imprecise bounding boxes. \\
\cite{son2022wearable} & Raw data supporting the conclusion & Addressing the need to ensure safe street crossing & Introducing a comprehensive wearable system for safe urban navigation, integrating real-time computer vision and prior maps. & Utilizes pre-built LiDAR maps, such as those publicly available and created for autonomous vehicles, supports user's preferred walking speed. & Dependent on specific crosswalk textures, lacks dynamic obstacle handling, may face instability with changing features, and outdoor noise interference. \\
\cite{hakim2019indoor} & $\bullet$ & Integration of navigation and object recognition. & Integration of navigation, object recognition, and low-cost sensors. & Integrated navigation and object recognition, low-cost sensors, efficient path planning, accurate real-time object identification in static scenarios. & Restricted to static simple environments. \\
\cite{goswami2023efficient} & $\bullet$ & Enhancing real-time global indoor localization using Semantic SLAM and a priori maps for GPS-deprived environments. & Implementing vector-based semantic extraction from floor plans, efficient particle filter localization, and leveraging loop closures for active semantic point cloud. & Real-time, accurate localization; integration of semantic information; using deep learning for enhanced robustness and efficiency. & Visual aliasing, limited semantic classes. \\
\cite{hao2022detect} & $\bullet$ & Addressing navigation challenges for BVI by developing a wearable solution for real-time guidance to target objects in unfamiliar settings. & Introducing a wearable navigation system with real-time object localization, visual SLAM, and trajectory estimation for efficient user guidance. & High accuracy, vision-based, real-time assistance, continuous object tracking, and portable system design. & Not explicitly mentioned. \\

\hline
\end{tabularx} 
     \begin{tablenotes}
    \item[2] \url{https://www.openicpsr.org/openicpsr/project/183714/version/V1/view} (Accessed on 3 May 2024)
    \end{tablenotes}
\end{threeparttable}
\end{table*}
\begin{table*}[t]
 \scriptsize
\centering
\caption{\textbf{A summary of perspectives and innovations in SLAM-based navigation solutions, with insights into limitations and advantages - Part III.}}
\label{tab:t_limitations_potential_solutions_partIII}
\setlength{\tabcolsep}{3pt}
\begin{threeparttable}
\begin{tabularx}{\textwidth}{p{20pt} >{\centering\arraybackslash}p{40pt} X X X X X }
\hline
\textbf{Ref.} & \textbf{Open-source} & \textbf{Research Issue} & \textbf{Contribution} & \textbf{Advantages}  & \textbf{Limitations} \\
\hline
\cite{kuribayashi2023pathfinder} & $\bullet$ & Addressing the challenge of navigation in unfamiliar indoor spaces by designing a map-less system, PathFinder & Developing a map-less navigation robot system incorporating sign recognition and intersection detection, using scenario-based participatory design with five blind participants. & Enhanced navigation confidence, provided key sign information, offered audio feedback, ensured safe, independent mobility by PathFinder & Limited study environments, assumed environments without steps or floor transitions, no empirical comparison with participants' regular aids, inability to recruit younger participants and insufficient number of guide dog users, possible positive bias from participants who have previously participated in studies, physical demand, Bluetooth connectivity issues, privacy concerns, inability to navigate in congested spaces, surrounding people may not recognize that users are disabled as they don’t use traditional aids. \\
\cite{agrawal2022novel} & $\bullet$ & Enhancing independent navigation and social seat selection with a perceptive robotic cane system. & Introducing a robotic cane with computer vision for navigation and social seat selection, featuring vibrotactile feedback and successful pilot validation. & Independent navigation, social norm-aware seat selection, intuitive vibrotactile feedback, and effective pilot-validated guidance. & Potential suboptimal chair detection, discrepancy between user preferences and performance. \\
\cite{rui2021multi} & $\bullet$ & Overcoming limitations of existing blind navigation methods by integrating multi-sensory feedback for comprehensive and intuitive mobility. &  Integration of YOLO and ORB-SLAM, enhanced by novel algorithms, provides reliable multi-sensory guidance & Enhanced accuracy, multi-sensory feedback, real-time object detection, dense navigation maps, and comprehensive guidance improve mobility and safety. & Not explicitly mentioned \\
\cite{ou2022indoor} & $\bullet$ & Addressing gap: aiding visually impaired in dynamic indoor navigation, obstacle detection, and social distancing. & Wearable RGB-D assistant system aiding indoor localization, mapping, and dynamic obstacle detection. & Dynamic object detection, enhanced obstacle avoidance, panoptic segmentation for scene understanding, robust tracking without additional models, and RGB-D sensor integration. & Imperfect panoptic segmentation, leading to errors in object recognition; computational complexity, influenced by the number of dynamic objects present in the scene.\\
\cite{jin2021wearable} & $\bullet$ & Addressing the gap: wearable device aiding visually impaired with indoor object manipulation tasks. & Hand-worn assistive device, RGBD-VIO method, effective human-device interface. & Enhanced pose estimation, depth information utilization, improved human-device interaction, effective object manipulation. & Not mentioned \\
\cite{xu2021multi} & $\bullet$ & Addressing the challenge of indoor navigation, integrating SLAM and deep learning enhances environmental perception. & Introducing multi-functional smart E-Glasses for enhanced indoor navigation and lightweight object detection. & Real-time navigation, high object detection precision, and robust SLAM integration. & Future enhancements aim to improve device comfort, portability, and size, addressing user concerns and enhancing overall usability. \\
\cite{lu2021personalized} & $\bullet$ & Addressing the gap in understanding between visually impaired individuals' perceptions and their actual surroundings for navigation technology. & Introducing personalized navigation system with object detection and description, and re-training models. & Personalized navigation, object detection, reduced time for finding destinations, and improved interaction. & Model re-training challenges, environmental complexity, and high computational cost. \\
\cite{kayukawa2020guiding} & $\bullet$ & Addressing the need for a guiding system to facilitate seamless walking in public spaces. & Introducing a comprehensive system aiding blind pedestrians by understanding nearby pedestrians' behavior. & Convenient suitcase design, accurate motion tracking, effective tactile interface for enhanced navigation. & Weight discomfort, space constraints, speed adjustment difficulties, and technological improvements for enhanced usability. \\
\cite{chen2020navigation} & $\bullet$ & Addressing the need for advanced navigation aid for the visually impaired using VSLAM technology. & Introducing a navigation aid system merging VSLAM with pre-established maps. & Not explicitly mentioned & Not explicitly mentioned \\
\cite{chen2020can} & $\bullet$ & Addressing the need for unified indoor navigation, integrating scene perception and visual localization. & Introducing a unified semantic visual localization system, enhancing obstacle avoidance and spatial awareness. & Real-time awareness, comprehensive understanding, and obstacle avoidance. & Restricted camera field of view and inconsistencies in semantic segmentation results impacting user confidence. \\
\cite{fusco2020indoor} & Upon publication & Addressing indoor navigation challenges through smartphone-based computer vision without new infrastructure. & Real-time app development, robust localization algorithm, and user-friendly navigation. & Cost-effective deployment, enhances usability, improves localization accuracy, promises full-featured wayfinding, and camera-agnostic navigation for ease of use. & Challenges in wide, open indoor spaces due to limitations in current approach, suggesting potential for augmented reality integration. \\

\hline
\end{tabularx} 
\end{threeparttable}
\end{table*}
\begin{table*}[t]
 \scriptsize
\centering
\caption{\textbf{A summary of perspectives and innovations in SLAM-based navigation solutions, with insights into limitations and advantages - Part IV.}}
\label{tab:t_limitations_potential_solutions_partIV}
\setlength{\tabcolsep}{3pt}
\begin{threeparttable}
\begin{tabularx}{\textwidth}{p{20pt} >{\centering\arraybackslash}p{40pt} X X X X X }
\hline
\textbf{Ref.} & \textbf{Open-source} & \textbf{Research Issue} & \textbf{Contribution} & \textbf{Advantages}  & \textbf{Limitations} \\
\hline
\cite{zhang2019human} & $\bullet$ & Addressing the indoor wayfinding problem & Introducing specialized robotic navigation aid, VIO method, guiding modes, and Human Intent Detection for enhanced navigation assistance. & Enhanced VIO method, two guiding modes, and automated mode selection. & Validation in larger spaces, user feedback incorporation, and operational restrictions. \\
\cite{zhao2019multi} & $\bullet$ & Addressing independent indoor corridor navigation through multi-sensor fusion and semantic mapping. & Semantic mapping, multi-sensor fusion, real-time performance enhancement & Enhanced corridor navigation, semantic mapping, landmark detection, real-time performance, multi-sensor fusion for improved navigation experience. & Object recognition limitations, arrow direction detection issue, training requirement for unknown objects. \\
\cite{weiss2020navigation} & \href{https://github.com/mweiss17/SEVN-data}{SEVN-data} \tnote{3} & Enhancing Reinforcement Learning environments to develop a navigation assistant tailored for the BVI community. & Developing a benchmark dataset and Reinforcement Learning training environment to advance navigation agent capabilities using real-world imagery and neural architecture. & SEVN offers realistic training with a rich, annotated dataset and a multi-modal fusion model for effective BVI navigation. & SEVN offers an extensible Reinforcement Learning environment, but improved model performance requires. \\
\cite{ramesh2018real} & $\bullet$ & Addressing real-time localization and navigation in indoor settings with a monocular camera, focusing on computational efficiency, user-friendly interfaces, and integrated algorithms. & Introducing a non-filter based visual SLAM with integrated object detection and distance-depth estimation algorithms, using a single monocular camera for BVI indoor navigation. & Utilizing a single camera and simpler SLAM algorithm for cost-effective, efficient real-time performance. & Not mentioned \\
\cite{eden2018indoor} & $\bullet$ & Enhancing indoor localization for visually impaired shoppers using text extraction, addressing gaps in traditional visual assistance systems. & Expanding SLAM algorithm for larger spaces using GIST/SURF features and navigating through text-rich environments. & Simplistic setup, no markers needed, and efficient real-time localization. & Operates only within length dimension of the aisle, limited by text density. \\
\cite{mulky2018autonomous} & Android APIs & Enabling safe autonomous navigation for elderly and visually impaired in crowded environments. & Design of an intelligent autonomous scooter with advanced sensor fusion, SLAM techniques, and hybrid mapping solutions. & Improved safety and autonomy, hybrid mapping for diverse environments, precise steering control. & Not mentioned \\
\cite{lalonde2018localizing} & $\bullet$ & Localizing pedestrians in crosswalks using visual odometry, addressing challenges in uniform textures and repetitive landmarks. & Introducing a prototype for localizing pedestrians in crosswalks using visual odometry. & Accurate on weakly textured surfaces, addressing scaling issues in monocular camera setups. & Initialization issues, tracking loss due to strong orientation variations, and challenges with oscillatory walking patterns. \\
\cite{zhang2017plane} & $\bullet$ & Improving indoor pose estimation accuracy for navigation aids using plane features in feature-sparse environments. & Introducing PAVIO method, utilizing plane features and factor graph optimization to improve pose estimation for indoor navigation. & Improved pose estimation accuracy and robustness, enhanced stability, accurate 3D mapping. & Not mentioned \\
\cite{chen2017ccny} & $\bullet$ & Addressing indoor navigation challenges to enhance mobility and independence. & Implementation of SmartCane with Google Tango for real-time indoor navigation and demonstration of its effectiveness. & Enhanced indoor navigation with real-time path planning, multimodal feedback, and an intuitive control panel interface. & Requiring further user evaluation to assess effectiveness \\

\cite{bai2017cloud} & $\bullet$ & Developing a cloud and vision-based system for safe navigation and detailed perception. & Integrating cloud computing with vision-based navigation, enhancing perception, and improving object recognition for blind individuals. & Detailed perception, real-person safety support, abundant surrounding information, and improved object recognition. & Requiring extensive vision-based mapping, struggles with similar scenarios, and needs improved scene parsing, currency validation, and object recognition. \\
\cite{endo2017indoor} & $\bullet$ & Addressing the need for specialized navigation systems for visually impaired individuals using SLAM technology for real-time guidance. & Introducing a wearable camera with LSD-SLAM for real-time positioning, obstacle detection, and route guidance. & Efficient calculation power, robust performance, accurate mapping, dynamic adaptation, real-time assistance. & Requires high-contrast environments for accurate mapping; low contrast may need external positioning solutions. \\
\cite{yang2022seeway} & $\bullet$ & Addressing the need for an innovative navigation system using vision-language model-based approach. & First BVI navigation system using spoken instructions, visual-language integration, and heuristic-based path planning for improved success. & Running on portable devices, provides BVI navigation without heavy labeling or 3D model reconstruction in complex indoor environments. & Navigation reliability drops for long distances. \\
\cite{shahani2023methods} & $\bullet$ & Addressing real-time navigation latency, safe route selection, and accurate obstacle detection. & Integration of Web of Things, predictive analytics, YOLOv4 Tiny, and SLAM for enhanced obstacle recognition and navigation. & Enhanced obstacle recognition, navigation, and safe route selection. & Requiring high-contrast environments; needing external updates when contrast is insufficient for accurate mapping. \\
\cite{yun2018design} & $\bullet$ & Addressing the need for efficient navigation solutions for visually impaired individuals and patients with lower body injuries indoors. & Developed a person carrier robot integrating Hector SLAM and Robust Salient Detection for safe navigation and obstacle avoidance. & Enhanced safety, improved mobility, and effective object detection for indoor navigation. & Sensitive to strong light, issues with reflective surfaces, and occasional collisions due to slow processing speed. \\

\hline
\end{tabularx} 
 \begin{tablenotes}
    \item[3] \url{https://github.com/mweiss17/SEVN-data} (Accessed on 12 May 2024)
    \end{tablenotes}
\end{threeparttable}
\end{table*}
%
\section{Result}
\label{results}
\begin{figure}[h]
    \centering
    \includegraphics[scale=0.4]{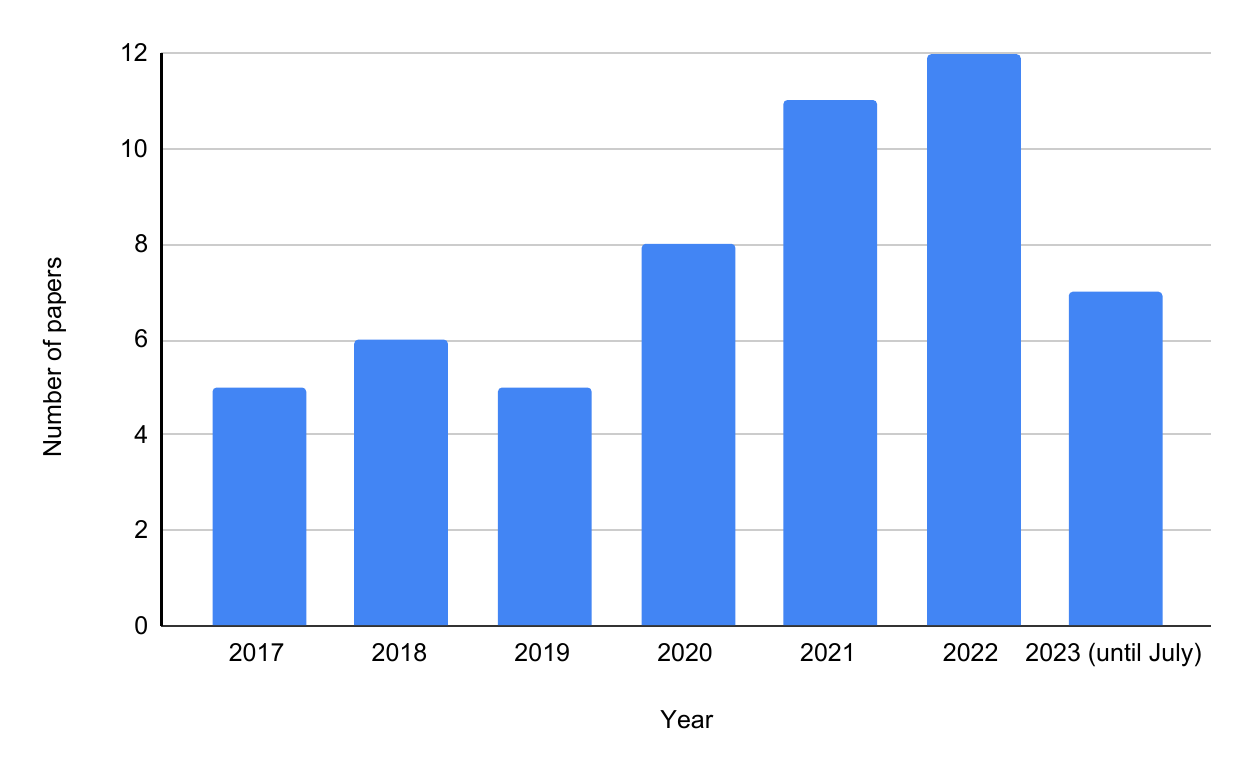}
    \caption{\textbf{Publications included in this review on SLAM-based BVI navigation, by year}}
    \label{distribution_of_papers}
\end{figure}

In this section, the findings of SLR are presented. Figure \ref{distribution_of_papers} shows the number of studies included in this review, which focused on BVI navigation using SLAM techniques. As shown in the figure, although only papers published in the first half of 2023 are included in this review, they constitute a substantial portion of the total. The growth in the number of studies in this domain suggests an advancement in SLAM techniques and an increase in their usage for developing navigation technologies for visually impaired individuals. 
This section is divided into four parts to answer the research questions. It discusses the types of SLAM techniques used to develop assistive technologies for visually impaired navigation, delves into the advantages and limitations of these techniques, highlights the challenging scenarios addressed, and presents the attributes of the SLAM technology that contribute to the enhancement of visually impaired navigation.

%
%
\subsection{RQ1. What localization and mapping approaches are used for the navigation of the visually impaired?}
Given the variety of SLAM systems designed with different sensors, applications, and scenarios, this section focuses specifically on reviewing the types of SLAM used for the navigation of the visually impaired. It is a key technology in robotics and computer vision and has the potential to assist visually impaired individuals with navigation. This can help visually impaired individuals provide real-time location information, maps, and spatial awareness. Among the 54 studies surveyed, three strategies were common, as shown in Table \ref{rq1_slam_type}. In this table, we use the exact terms mentioned in the literature for the localization and mapping techniques.

To further understand the technical features employed in these solutions, the detailed information is presented in Tables \ref{tab:core_technical_features_partI}–\ref{tab:core_technical_features_partIII}. These tables focus on the localization and mapping components of the assistive system, specifically highlighting the sensor types, computing resources, and application of machine learning-based methods. By examining these features, we can gain deeper insight into how these systems are structured and the diverse technologies utilized to achieve accurate and efficient SLAM for assistive navigation. It is important to note that this information relates only to the localization and mapping components of the assistive navigation solutions. Details of the entire system are provided in Tables \ref{tab:t_prototype_information_wearable_partI}–\ref{tab:t_prototype_information_misc}.

This section is divided into three subsections, where we discuss the localization and mapping approaches, the sensor types used for these approaches, and the computing resources required to perform these approaches.

\subsubsection{Approaches}
The majority of studies have leveraged established SLAM techniques, such as ORB-SLAM, while some studies have developed new solutions tailored to their needs. For example, \cite{jin2020combining} proposed visual simultaneous localization and mapping for the moving-person tracking (VSLAMMPT) method. The proposed method was designed to assist people with disabilities, particularly visually impaired individuals, in navigating indoor environments. Additionally, various studies have used the SLAM components of existing frameworks, such as ARCore and ZED camera SLAM.

It is worth noting that several studies have employed VIO and SLAM as the core components in their proposed systems, whereas others have employed them to enhance the robustness of localization \cite{cheng2021hierarchical}, for comparison with alternative localization approaches \cite{lu2021assistive}, or to develop new localization methods \cite{crabb2023lightweight}. For example, \cite{crabb2023lightweight} presented a novel localization approach specifically designed for individuals with visual impairment. This method combines visual landmark identification, VIO, and spatial constraints derived from a two-dimensional (2D) floor plan.

SLAM methodologies can be categorized into feature-based, direct, and optical-flow techniques. Feature-based methods extract and describe feature points in an image, which are then matched across different images for tracking and mapping. On the other hand, direct methods directly calculate the luminosity changes of pixel blocks. Optical flow methods utilize the optical flow changes in feature points, pixel gradient points, or the entire image to track and map the environment \cite{chen2021wearable}.

SLAM algorithms are further categorized into optimization- and filtering-based methods, each with distinct approaches to map creation and agent localization. Optimization-based, often referred to as Graph SLAM, treats the problem as a large optimization task, where the goal is to find the set of poses and landmark positions that best explain the observed sensor measurements. This is typically achieved by constructing a graph in which nodes represent  agent poses or landmarks and edges represent constraints or observations between them. The solution is determined by minimizing the global cost function, which represents the error between the predicted and actual measurements, using nonlinear optimization techniques. On the other hand, filtering-based SLAM uses recursive Bayesian filters, such as the Extended Kalman Filter (EKF) or Particle Filter, to incrementally update the map and the  agent's position as new sensor data arrive.

Systems such as ORB-SLAM and RTAB-Map are feature- and optimization-based, employing features and graph optimization for mapping and localization. Conversely, LSD-SLAM and DSO are examples of direct and optimization-based SLAM. Some systems, such as Semantic SLAM, may adopt either approach, depending on their implementation. It is important to note that the method and type of SLAMs are not directly mentioned in all papers, so the information provided here is a general categorization based on common practices within each category.

The reviewed studies demonstrate the versatility of SLAM in various navigation scenarios. The specific implementation and aspects addressed in each study varied depending on the application. SLAM can be employed in environments that lack a map and can dynamically create it while navigating. This involves simultaneous map creation and localization within an environment. Alternatively, SLAM can be employed to generate maps for subsequent navigation. In this case, SLAM first builds a map of the environment and then the map is utilized during navigation. Studies have also used SLAM odometry for navigational tracking. Odometry provides a continuous estimate of the position and orientation of an agent based on the sensor readings.

%
%
\begin{table*}[h]
\scriptsize
\centering
\caption{\textbf{Localization and mapping approaches used for visually impaired navigation, with "NA" denoting data not available. }}
\label{rq1_slam_type}
\setlength{\tabcolsep}{3pt}
\begin{tabularx}{\textwidth}{p{75pt} p{75pt} p{100pt} p{40pt} p{100pt}}
\hline
\multicolumn{5}{ c }{ \textbf{Widely applicable techniques}} \\
\hline
\textbf{Technique} & \textbf{Type} & \textbf{Method} & \textbf{Sensor type} & \textbf{Reference(s)} \\
\hline
Cartographer & Scan matching & Optimization-based & LiDAR & \cite{kuribayashi2023pathfinder, zhang2023research, zhou2022multi, kayukawa2020guiding} \\
Hector SLAM & Scan matching & Optimization-based & LiDAR & \cite{hakim2021indoor, yun2018design, hakim2019indoor} \\ 
LiDAR SLAM & Scan matching & \textcolor{violet}{Rao-Blackwellized particle filter} & LiDAR & \cite{albogamy2021sravip, lu2021assistive, slade2021multimodal} \\
FastSLAM \cite{montemerlo2002fastslam} & Feature-based & \textcolor{violet}{Particle filter-based} & LiDAR & \cite{son2022wearable}: mapping \\
Kinect Fusion & Feature-based & Optimization-based & \textcolor{my_DarkGreen}{Visual} & \cite{song2023mixture} \\
LSD-SLAM & \textcolor{blue}{Direct} & Optimization-based & \textcolor{my_DarkGreen}{Visual} &  \cite{endo2017indoor, shahani2023methods} \\
\hline
\end{tabularx}
\end{table*}

Our analysis, underscored by the classifications in Table \ref{rq1_slam_type}, indicates a strong preference for feature-based and optimization-based SLAM approaches for visually impaired navigation. This preference is likely due to the robustness and efficiency of these methods in processing visual data, which is key for real-time assistive navigation. 

\begin{figure*}
    \centering
    \includegraphics[scale=.5]{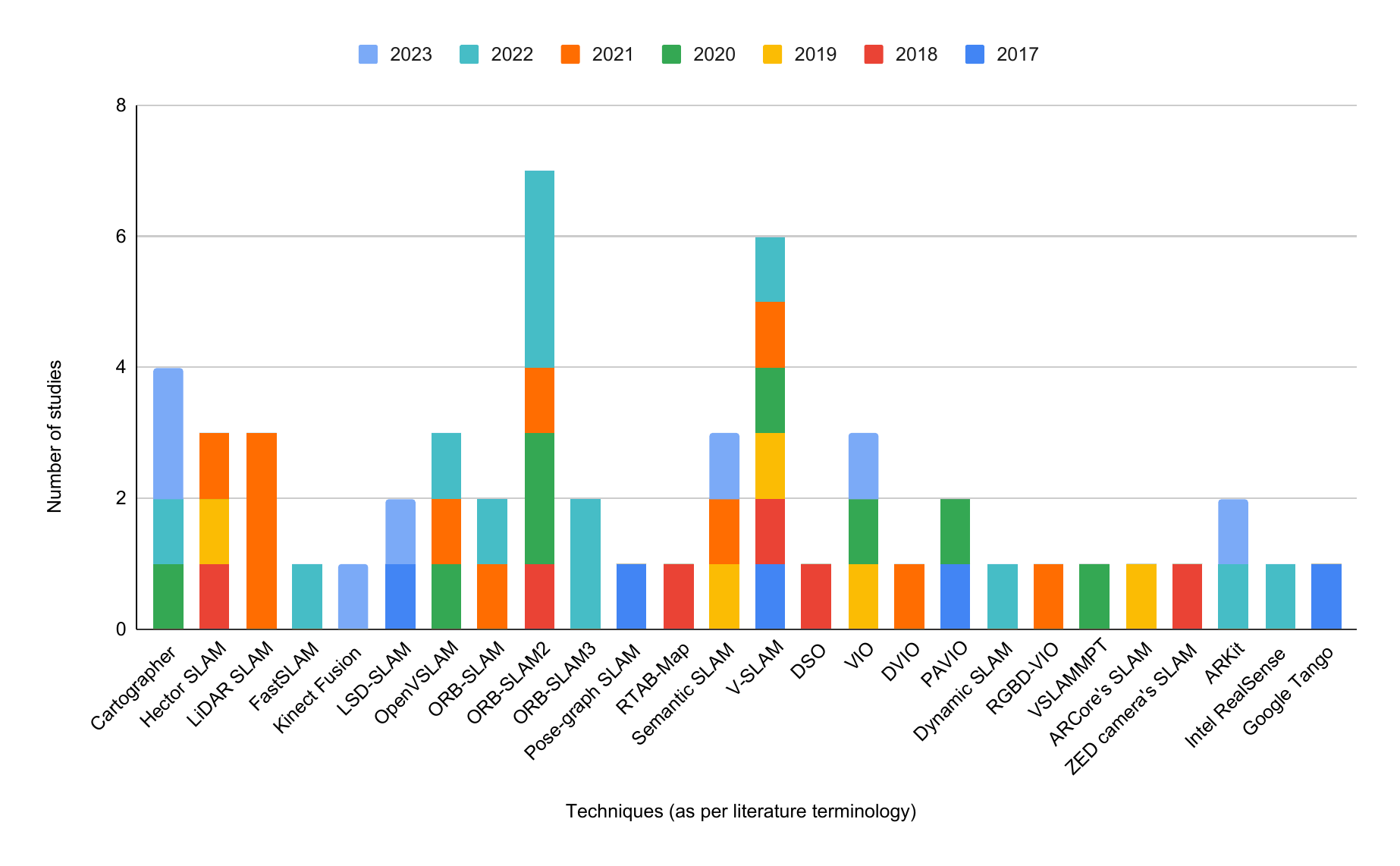}
    \caption{\textbf{Evolution and adoption of localization and mapping techniques in BVI navigation systems over time.}}
    \label{SLAM_over_time}
\end{figure*}

Figure \ref{SLAM_over_time} provides insight into the use of various localization and mapping techniques for visually impaired navigation from 2017 to the date when SLR was conducted (July 2023). This figure illustrates that visual techniques has consistently been used across all years. Although many other techniques also operate based on visual data, we mentioned each of these techniques, as indicated in the referenced studies. 

The utilization of semantic SLAM and Cartographer SLAM signifies a recent trend towards leveraging advanced spatial understanding and mapping capabilities for visually impaired navigation. Semantic SLAM incorporates higher-level scene interpretation and enhances users’ contextual awareness. On the other hand, Cartographer SLAM provides SLAM in 2D and 3D across various platforms and sensor configurations, offering innovative solutions to tackle the diverse challenges associated with BVI navigation.

ORB-SLAM algorithms, including ORB-SLAM (published in 2015), ORB-SLAM2 (published in 2017), and ORB-SLAM3 (published in 2021), have gained popularity because of their robustness and performance. This can be attributed to its efficient feature extraction and matching techniques, making it well suited for real-time navigation applications.

Customized techniques have been developed to meet specific needs. This trend indicates that researchers have adjusted the SLAM techniques to better match the specific requirements of their intended applications. This suggests closer integration of SLAM with domain-specific needs.

%
%
\subsubsection{Sensor type}
\begin{figure}[h]
    \centering
    \includegraphics[scale=0.6]{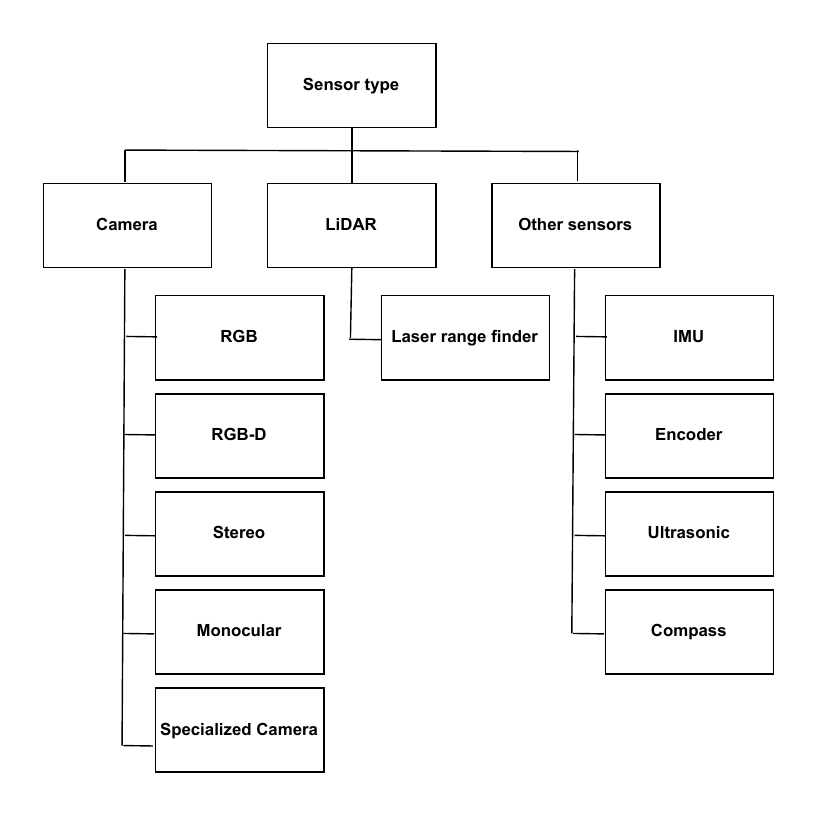}
    \caption{\textbf{Overview of sensor types in studies under review}}
    \label{sensor_type}
\end{figure}

The sensors employed in SLAM solutions for BVI navigation are diverse and include various types of cameras, LiDAR, IMU, and other specialized sensors. 
As shown in Figure \ref{sensor_type}, we categorized the sensors into three types: cameras, LiDAR, and other sensors. 

\input{texs/tables/t_core_technical_features_partI}
\input{texs/tables/t_core_technical_features_partII}
\input{texs/tables/t_core_technical_features_partIII}

\textbf{Camera}
The common use of visual sensors in SLAM techniques can be attributed to advances in computer vision and image processing, which enhance navigation capabilities by providing rich environmental information. This makes visual-sensor-based SLAM techniques the most commonly used in the implementation of assistive technologies for the BVI people, offering a cost-effective, versatile, and accurate solution for real-time navigation and spatial awareness. The literature under review used the following types of cameras: RGB, RGB-D, stereo, monocular, and other specialized cameras.

RGB cameras are widely used due to their ability to capture rich color information, which is beneficial for visual odometry and object recognition. They are cost effective and widely available, making them a popular choice for the development of accessible navigation aids.

RGB-D cameras provide both color and depth information, enabling more accurate mapping and localization. Depth information helps in understanding the 3D structure of the environment.

Stereo cameras also provide depth perception through two slightly offset lenses that simulate binocular vision. They are effective in capturing detailed depth information and are useful in applications where precise depth estimation is required.

Monocular cameras are simpler than stereo and RGB-D cameras. They rely on visual odometry and other techniques to estimate depth and motion, making them lightweight and suitable for mobile applications.

Specialized cameras, including fisheye, 3D time-of-flight, and wide-angle cameras, provide specialized capabilities, such as a wide field of view or precise depth measurement, which can enhance the SLAM performance in specific scenarios.

\textbf{LiDAR}
LiDAR sensors are highly accurate in measuring distances and are effective in creating detailed 3D maps of the environment. Studies use LiDAR alone to build a map of the environment or in combination with other sensors such as IMU and cameras to enhance the robustness and accuracy of SLAM systems.

\textbf{Other sensors}
Various studies combined different types of sensors to leverage the strengths of each type and provide more robust and reliable navigation solutions. For example, integrating an IMU with a camera helps achieve better motion tracking and stability. This trend towards integrating multiple sensors highlights increasing efforts to enhance the robustness and reliability of SLAM solutions.

Table \ref{tab:t_sensor_type_classification} provides a detailed breakdown of the sensor types employed in the reviewed studies, highlighting the prevalence of different sensor modalities and their combinations in SLAM-based assistive technologies for BVI individuals.
\input{texs/tables/t_sensor_type_classification}

The reviewed papers show a clear preference for RGB-D cameras, indicating their effectiveness in providing both the visual and depth information necessary for accurate SLAM applications. The use of LiDAR is important in applications that require precise mapping. Over the years, there has been a noticeable trend towards integrating multiple sensors and combining their strengths to achieve more robust and reliable SLAM solutions for visually impaired navigation. The integration of machine learning-based techniques with SLAM systems is particularly prevalent in solutions that utilize RGB-D cameras. This highlights the effectiveness of combining this type of data with advanced AI algorithms. This trend is likely to continue as technology advances, offering more sophisticated and adaptable solutions in complex and dynamic environments.

%
%
\subsubsection{Computing resource}
To process data and run localization and mapping algorithms, the reviewed studies adopted two classes of computational resources: local and remote. Local computations are performed in situ on devices, such as smartphones, tablets, laptops, portable microcontrollers, and UP board computers. In some cases, algorithms were applied on PCs. Table \ref{tab:computing_resources} categorizes the computing resources used in the reviewed studies for localization and mapping tasks. Information regarding the computing resources for the entire navigation assistive system is detailed in Tables \ref{tab:t_prototype_information_handheld_partI}-\ref{tab:t_prototype_information_misc}
\input{texs/tables/t_computing_resources}
\\
\\
\paragraph{\textbf{Local computing resources}}

\textbf{Smartphones:} Smartphone are widely used as communication gadgets, and their technology continues to grow to the point that it is possible for smartphones to implement functional navigation systems. Because smartphones integrate diverse sensors such as IMU, GPS, and cameras, they can be used as a convenient tool for collecting environmental information. In addition, their computational power can be exploited to perform various navigation operations. For example, the system proposed by \cite{bai2019wearable} implemented all algorithms relevant to data acquisition, ground segmentation, moving direction search, global path planning, indoor and outdoor localization, and object detection on a smartphone and achieved real-time performance. Without an additional depth sensor, \cite{zhang2019arcore} took advantage of an ARCore-supported smartphone to track pose and to build a map of the surroundings in real time. However, despite the significant advantages of smartphones, such as their small size, low weight, easy portability, and low cost, their computing power is not sufficient for some approaches.

\textbf{Laptops and PCs:} Some of the reviewed approaches perform all or part of the required calculations locally on a portable computer, such as a laptop. Despite higher computing power compared to a smartphone, and greater security compared to remote computational resources, the laptop’s heavier weight and large size are considered major disadvantages, especially during long trips. PCs provide even higher computing power, which is essential for complex SLAM operations; however, they lack portability.

\textbf{Embedded systems and microcontrollers:} Embedded systems such as Nvidia Jetson boards, Raspberry Pi, and UP boards provide a balance between computational power and portability. They are commonly used in the reviewed studies for performing SLAM operations locally. For instance, \cite{song2023mixture} utilized a Hololens2 device with a GPU for real-time 3D reconstruction, whereas \cite{hou2022knowledge} used a Jetson AGX Xavier for ORB-SLAM2 and dense mapping. Raspberry Pi devices are also popular due to their low cost and sufficient computing power for many SLAM tasks \cite{hakim2021indoor, slade2021multimodal}.
\\
\\
\paragraph{\textbf{Remote computing resources}}

An alternative solution is to transfer all or part of the calculations to the remote computing resources. To reduce local computing costs, \cite{li2022sensing} adopted an embedded computer and a remote server. In the proposed vision-based assistance system, before transferring the input images to the server, the images were time-stamped and encrypted on an embedded computer. The remote server was equipped with a CPU and GPU to run parallel ORB-SLAM2 and artificial intelligence algorithms for indoor navigation, object detection, face recognition, and scene text recognition. Experiments confirmed that the use of remote servers under a smooth network connection, such as 4G or WiFi, can meet the computational requirements of the proposed system. However, although the high computing power of remote servers is considered a significant advantage, constant Internet access over a secure connection is required. Moreover, the performance of the entire system would be affected by the network condition.

%
%
\subsection{RQ2. What are the advantages and limitations of SLAM techniques for BVI navigation?}
Although SLAM techniques offer significant benefits for navigation across various applications, their use in improving mobility for BVI individuals presents unique considerations. In general applications, the advantages of SLAM include accurate real-time mapping and localization, adaptability to unknown environments, and the ability to function without an external infrastructure. These advantages are essential when applied to BVI navigation. For instance, the ability to provide real-time accurate spatial information is crucial for safe navigation and obstacle avoidance.

Conversely, some limitations of SLAM in general applications, such as computational complexity and sensor dependencies, pose significant challenges in the BVI context. The need for lightweight, portable devices with a long battery life is critical for BVI users, who rely on these systems for extended periods. In addition, robust performance in diverse environments, including crowded spaces and varying lighting conditions, is vital for effective BVI navigation. Unique to BVI applications is the need to translate complex spatial data into intuitive, non-visual feedback. Furthermore, the integration of semantic information to provide context about the environment (e.g., identifying doors, stairs, or pedestrian crossings) in order to interact with the environments during navigation is particularly valuable for BVI users, but may be less critical in other SLAM applications.
In the following subsections, we explore in detail the advantages and limitations of SLAM techniques when applied to BVI navigation.

\subsubsection{Advantages}
Unlike many localization approaches, such as RFID- or GPS-based methods, which require infrastructure setup, SLAM does not depend on pre-existing infrastructure. It operates autonomously by creating maps and understanding the surroundings in real-time.
SLAM relies on data captured by sensors already present on many mobile devices, such as smartphones, and offers a cost-effective solution for accurate localization and mapping without the need for additional hardware or subscription services.

One of the most important advantages of SLAM is its potential for real-time positioning, which determines the agent’s current location and orientation. Systems leveraging ORB-SLAM, for instance, excel in pose estimation by integrating various data types, including visual, inertial, and depth information, thus enhancing accuracy beyond conventional methods \cite{xie2022multi}. This feature is pivotal not only for effective navigation but also for obstacle detection and avoidance, ensuring the safety and confidence of visually impaired users as they navigate through immediate environments \cite{xie2022multi, rui2021multi}.

SLAM's ability to reuse and update maps incrementally allows for a high degree of environmental adaptation. Its capacity to relocalize within prebuilt maps or expand them as necessary ensures that users can rely on updated information for navigation \cite{zhang2019arcore, bai2018virtual}. This adaptability is further enhanced by the capability of the system to handle dynamic environments, making it invaluable for visually impaired users who require real-time path adjustments in response to moving obstacles \cite{jin2020combining, ou2022indoor}.

Detailed environmental mapping facilitated by SLAM, ranging from two-dimensional layouts to complex 3D geometric and semantic maps, provides comprehensive spatial understanding \cite{zhang2023research, slade2021multimodal, liu2020indoor}. Environmental awareness is critical for path planning and collision avoidance. Furthermore, the integration of semantic mapping enriches spatial understanding by adding contextually rich information to maps, thereby facilitating more informed decision-making and interaction with the environment \cite{chen2021wearable, zhao2019multi}.

The integration of different types of sensors and technologies with SLAM significantly expands the scope of its application. By integrating techniques such as object detection algorithms or combining RGB-D and IMU sensor data, SLAM systems achieve a multilayered perception of the environment \cite{xie2022multi, goswami2023efficient}. Sensor fusion enhances a system's ability to detect and classify objects, accurately navigate, and handle dynamic elements within the environment, thereby offering a more holistic assistive solution \cite{goswami2023efficient}.

The cost-effectiveness of SLAM-based solutions attributed to their reliance on widely available low-cost sensors makes this technology particularly interesting. Systems employing monocular cameras or wearable RGB-D cameras exemplify how SLAM can be implemented in a cost-effective manner without compromising functionality, thus making advanced navigational aids accessible to more users \cite{hakim2021indoor, hao2022detect}.
The advantages of SLAM, derived from the literature, are listed in Table~\ref{rq2_advantages}.

\input{texs/tables/t11_RQ2_SLAM_advantages}

\subsubsection{Limitations}
Although SLAM technologies show great potential for improving navigation aids for the visually impaired, they are not without their limitations. These limitations can significantly impact the effectiveness and reliability of SLAM-based assistive systems.

A notable challenge is the computational complexity and the associated demand for system resources. The implementation of advanced SLAM algorithms and the integration of deep learning frameworks for semantic understanding introduce significant computational overhead \cite{li2022sensing, liu2020indoor}. This complexity can compromise the real-time performance, which is crucial for assistive navigation. The need for appropriate hardware to process high-resolution data further underscores this limitation, potentially restricting the deployment of SLAM-based systems \cite{rui2021multi}.

The effectiveness of SLAM is dependent on its environmental characteristics. Accurate mapping and localization depend on the presence of distinct geometric features. In environments lacking such features or dynamically changing settings, SLAM systems may struggle to maintain accurate localization, thereby leading to navigation errors \cite{zhang2017indoor}. This limitation is particularly evident in feature-poor areas such as long corridors or spaces with uniform surfaces, where loss of localization can occur \cite{hou2022knowledge}.

Another critical limitation is the dependence on initial data or pre-existing maps. Some SLAM systems require sighted individuals to pre-map the environment, which can limit the flexibility and immediate usability of unmapped or altered spaces \cite{zhang2019arcore, bai2018virtual}. This reliance on prior mapping can be a significant hurdle for deploying SLAM-based navigation aids in diverse and changing environments \cite{son2022wearable, chen2020navigation}.

Drifting errors present a substantial challenge for maintaining the long-term accuracy of SLAM systems. Over time, small inaccuracies can accumulate, leading to significant deviations from the true trajectory, which can disorient users and compromise the navigation safety \cite{ahmetovic2023sonification}. In addition, the ineffectiveness of some SLAM systems for generating dense navigation maps limits their utility in providing the detailed guidance required for visually impaired navigation, necessitating further algorithmic enhancements \cite{xie2022multi}.

The performance of SLAM in dynamic environments, characterized by moving obstacles and changing conditions, remains a critical concern. Systems may fail to adapt quickly to such changes, leading to potential navigation errors and safety risks for visually impaired users \cite{lu2021assistive}.

Some SLAM applications require external calibration or setup, such as placement of calibration boards in specific environments. This requirement can limit the spontaneity and ease of use of SLAM-based navigation aids because it imposes additional  constraints \cite{lalonde2018localizing}. 
Table \ref{rq2_limitations} outlines the overall limitations of SLAM, derived from the publications under review.

\input{texs/tables/t12_RQ2_SLAM_limitations}

%
%
\subsection{RQ3. What challenging situations have been addressed?}
This section explores various challenging situations addressed by SLAM-based navigation-assistive systems for BVI individuals. We categorized these challenges into two main groups: those relevant to environmental complexities and those related to the sensors used for receiving environmental data. Additionally, we discuss practical challenges and considerations that impact the usability and adoption of these systems.

\subsubsection{Technical and methodological challenges}
Optimal pathfinding, perception of surroundings, and obstacle avoidance are crucial for navigation. Precise localization of a visually impaired user within the environment is essential for the effective operation of these functions. Our surveyed papers addressed the localization and mapping problems using various techniques. Some of these studies explored other navigation-related challenges. We categorized these challenges into two groups: those relevant to environmental complexities, and those related to the sensors used to receive environmental data. Dynamic obstacles and crowded spaces constitute the challenges in the first group, whereas challenges related to changing lighting conditions and the rapid motion of users that results in motion blur fall into the second group. Table \ref{rq3_challenges} lists studies that investigated these challenges through the integration of SLAM with other approaches. In the following section, we discuss the studies that address these challenges.

%
%
\begin{table}
\footnotesize
\centering
\caption{\textbf{Challenges addressed in reviewed studies using SLAM techniques}}
\label{rq3_challenges}
\setlength{\tabcolsep}{3pt}
\begin{tabular}{p{50pt} p{55pt} p{110pt}}
\hline
\textbf{Related to} & \textbf{Challenges} & \textbf{Reference(s)} \\
\hline
\multirow{2}{*}{\textbf{Environment}} & Crowded places & \cite{mulky2018autonomous, zhang2023research, qiu2022egocentric, lu2021assistive, kayukawa2020guiding} \\

& Dynamic object & \cite{zhao2019multi, cheng2021hierarchical, zhang2023research, ou2022indoor, lu2021assistive, jin2020combining, endo2017indoor, shahani2023methods} \\
\hline
\multirow{2}{*}{\textbf{Sensor}} & Fast motion & \cite{liu2020indoor} \\

& Illumination & \cite{zhou2022multi, cheng2021hierarchical, liu2020indoor, bai2019wearable, salihappearance} \\ 
\hline
\end{tabular}
\end{table}

%
\paragraph{Environmental complexities}

%
%
\textbf{Crowded scenarios}
Navigating crowded environments presents significant challenges for the visually impaired, leading to increased collision risks and difficulties in maintaining personal spaces. The absence of visual information makes it difficult to measure distance, perceive crowd density, and locate landmarks or places of interest.

Navigation in crowded environments also poses challenges for assistive technologies. For example, in assistive systems that operate based on SLAM, the presence of numerous dynamic elements, such as moving individuals and objects, introduces ambiguity into feature detection and tracking, leading to difficulties in accurately estimating the pose of the user and structure of the environment. The dynamic nature of crowds also hinders loop closure detection, disrupts map consistency, and contributes to drift. Moreover, the lack of distinct visual landmarks in crowded scenes represents a reliable localization challenge, which potentially reduces the robustness and accuracy of the overall SLAM system. Addressing these challenges requires the development of new approaches to address the complexity of such environments effectively. Several studies have investigated this issue. 

\cite{zhang2023research} presented a guide mobile robot engineered for the complexities of navigating different environments while considering dynamic objects and human presence. The robot could handle crowded environments with multiple dynamic objects. To accomplish this, the robot leveraged a spatial risk map, which is a tool that evaluates potential object-occupied spaces, to chart a path that effectively minimizes disruptions. This study presents experiments in which a robot successfully guided a user through the passage of multiple objects and people. The research used Cartography SLAM for off-line mapping. It's important to note that this paper did not address the dynamic environment through SLAM, but rather used SLAM solely as a tool to pre-build the map of the environment. 

In another study, \cite{qiu2022egocentric} introduced an egocentric human trajectory forecasting model that was designed for navigation in crowded environments. The model predicts the path of the sensor wearer using their past trajectories, nearby pedestrian trajectories, scene semantic and depth data. The authors collected an egocentric human trajectory forecasting dataset. As they could not use GPS or motion capture systems for recording the trajectory, they used ORB-SLAM3 to obtain the ground-truth sensor wearer trajectory. The trajectories obtained using ORB-SLAM3 were used to train the egocentric human trajectory forecasting model. It is important to note that this study, like the previous one, did not handle the crowded environment through SLAM itself, but rather used SLAM as a tool for obtaining  the trajectory ground-truth.

In addition, \cite{lu2021assistive} addressed challenges in crowded environments using a combination of SLAM and Ultra-Wideband (UWB) positioning. However, the SLAM algorithm was found to be less effective in environments with dynamic obstacles such as pedestrians. The algorithm finds features of dynamic obstacles moving along with the robot as the assistive device, and thus, it was misled that the robot did not move at all. However, using UWB positioning mitigates this issue.  

In \cite{kayukawa2020guiding}, the method addressed crowded environments by recognizing and predicting people's behavior while anticipating the collision risk. The system advises users to adjust their walking speed (on-path mode) or to choose alternative routes (off-path mode). This involves comparing the 3D point cloud map to real-time LiDAR and IMU sensor data. The system then predicts the future position and velocity of the user in order to avoid collisions.  

An intelligent autonomous scooter was developed in \cite{mulky2018autonomous} for navigating environments with small safety margins and highly dynamic pedestrian traffic such as sidewalks with numerous obstacles and pedestrians. The authors proposed a hybrid mapping solution that combines far-field and near-field mapping to navigate through dynamic environments. This approach utilizes sensor fusion to adapt dynamically to complex and cluttered environments. However, it should be noted that this study conducted system tests in a completely static environment without moving objects, such as pedestrians. Furthermore, the RTAB-MAP SLAM system was used in this study without any adaptation to dynamic environments.

%
%
\textbf{Dynamic objects}
Several publications have not directly addressed the challenge of crowded environments, and have only focused on dealing with the presence of dynamic objects within the scene. In the system proposed by \cite{ou2022indoor}, dynamic objects can be identified, and average depth information can be provided to the user. When a dynamic object belongs to a predefined class such as a person, it can also be tracked between frames in the SLAM pipeline. The system is capable of identifying and tracking dynamic objects after ego-motion estimation to obtain average depth information. Subsequently, it can estimate the poses and speeds of these tracked dynamic objects and relay this information to the users through acoustic feedback. Depth information helps users maintain social distancing in public indoor environments such as shopping malls.  

To address the challenge of dynamic objects, \cite{jin2020combining} proposed a new method called visual simultaneous localization and mapping for moving person tracking (VSLAMMPT). This method was designed to handle dynamic environments in which objects constantly move. The system also uses expected error reduction with active-semi-supervised learning (EER–ASSL)-based person detection to eliminate noisy samples in dynamic environments. This aids in accurate detection and avoidance of dynamic obstacles.  

\cite{zhao2019multi} utilized YOLOv3 to detect common objects in a corridor, including people, which were identified as obstacles. The system sends information about obstacles to users every five seconds when the distance between the user and obstacle is less than 10 m. For example, it may notify the user, "A person is located 2.8 meters ahead."

%
%
\paragraph{Sensor-related challenges}
\textbf{Changes in lighting condition}
Lighting changes pose a hurdle to visual SLAM systems. Illumination variations alter the visual features, interfere with accurate detection and matching across frames, impact pose estimation, and map-building robustness. SLAM relies on distinctive features for operation; however, lighting changes introduce ambiguities, noise, and errors, which affect the accuracy. Overcoming this challenge requires robust algorithms for dynamic lighting to ensure stable and precise localization and mapping.  

The method proposed in \cite{cheng2021hierarchical} tackles the challenge of illumination changes using a deep descriptor network called a Dual Desc, which is designed to be robust against various appearance variations including illuminance changes. The network used multimodal images (RGB, Infrared, and Depth) to generate robust attentive global descriptors and local features. These descriptors were used to retrieve coarse candidates from query images, and 2D local features, along with a 3D sparse point cloud, were used for geometric verification to select the optimal results from the retrieved candidates. The authors mentioned that their dataset included images captured at different times of the day, which resulted in illumination changes between the query and database images. Despite these changes, the proposed method achieved satisfactory localization results.  

The authors of \cite{liu2020indoor} evaluated the influence of lighting conditions on the performance of their novel localization method. The authors captured training images during the day and test images at night and simulated changes in lighting conditions by switching some of the lights off in locations without windows. The results showed that changes in the lighting conditions had a minor impact on the proposed method.  

\cite{zhou2022multi} mentioned that the proposed localization scheme was verified in a typical office building environment with dramatically changing lighting conditions throughout the day; however, it does not provide detailed results or discussion on how changing lighting conditions affect the performance of the system.  

The method proposed by \cite{salihappearance} addressed changes in illumination as a challenge using the COLD and IDOL datasets, which were recorded under different weather and illumination conditions (cloudy, night, and sunny) using different mobile platforms and camera setups. These datasets were used to evaluate the strength of the localization and recognition algorithms with respect to the variations caused by human activities and changes in illumination conditions. The study also mentioned the use of Histogram of Oriented Gradients (HOG) for feature extraction, which provides preferable invariant results for lighting and shadowing.

%
%
\textbf{Motion blur}
Blurred images in visual SLAM can lead to inaccuracies in feature detection and matching, causing issues with pose estimation, map building, loop closure, and visual odometry. These inaccuracies can also impact depth measurements and map quality. To address this issue, strategies such as using high-frame-rate sensors, incorporating IMUs, and employing motion-deblurring techniques can be employed to improve the accuracy of localization and mapping in SLAM systems.  

Motion blur can be caused by fast or sudden movements of the user during navigation, which can affect localization performance. \cite{liu2020indoor} studied this challenge and evaluated the robustness of the localization methods. They captured 2316 blurred images on the testing day. The results show that the proposed method performed poorly in this experiment, indicating that fast motion or sudden changes in user movement can pose a challenge to the system. The reason for this poor performance is that the object detection scores did not exceed the threshold during the experiment.

\subsubsection{Practical challenges and considerations}
In addition to the technical and methodological aspects, we recognized the importance of practical challenges and considerations that can significantly affect the usability and adoption of SLAM-based assistive systems. Therefore, we included an evaluation of the practical challenges and operational efficiency, as summarized in Tables \ref{tab:practical_challenges_partI} and \ref{tab:practical_challenges_partII}. The information in these tables has either been directly extracted from the article or can be easily inferred from the article's text. These tables provide information on user-friendliness, cost-efficiency, weight, comfort for extended use, adjustable fit, fatigue mitigation, and portability of the assistive tools described in the reviewed studies. For instance, while smartphones and lightweight devices such as eyeglasses-mounted sensors \cite{bai2019wearable} and ARCore-supported smartphones with haptic gloves \cite{zhang2019arcore} are generally well received because of their high portability and ease of use, heavier devices such as guiding robots \cite{lu2021assistive} and rolling suitcase-shaped device \cite{kayukawa2020guiding} are noted to cause user fatigue over extended periods. The augmented cane \cite{slade2021multimodal}, although found to improve confidence and workload for novice and expert users, also faced usability challenges owing to its weight. Clear instructions and easy learning curves, as seen in electronic glasses with haptic modules \cite{li2022sensing}, play a significant role in enhancing user satisfaction. However, the cost efficiency of these technologies varies, with some solutions being more affordable and accessible.

\input{texs/tables/t_practical_challenges_partI}
\input{texs/tables/t_practical_challenges_partII}

%
%
\subsection{RQ4. How the proposed solution is expected to enhance mobility and navigation for visually impaired?}
This section discusses how the approaches proposed by the studies included in our SLR have the potential to improve navigation for BVI people. These studies focused on diverse attributes, such as accurate pose estimation, semantic mapping, sensor fusion, and algorithmic innovations, to improve the quality of BVI navigation. Table \ref{rq4_mobility_improvement} presents the categorization of attributes that contribute to enhancing the mobility and navigation of visually impaired individuals.

To understand the impact of these solutions further, we examined their effectiveness in real-world scenarios. Various localization and mapping techniques have been assessed on the basis of their accuracy, robustness, consideration of dynamic objects, and running time. This evaluation provides insight into the performance of these techniques in practical environments.

In addition, we considered user-based evaluations to gauge user satisfaction and the practical applicability of the proposed system. These evaluations include feedback from actual users, which is crucial for understanding real-world usability and acceptance of assistive technologies.

Furthermore, we provide a detailed overview of the components and technologies used in assistive navigation systems. This helps to understand the practical implementations and innovations proposed by the studies. By examining the system prototypes, we gained insights into the design and functionality of assistive solutions beyond the localization and mapping components. This offers a comprehensive view of how these technologies enhance the mobility and navigation of the visually impaired.

\subsubsection{Attributes enhancing mobility and navigation}
SLAM technology is primarily used to provide precise localization, which is critical for assistive navigation systems. Precise localization provides accurate information regarding a user's position in the environment in which the user navigates. This accuracy enables the system to offer feedback on obstacles, pathways, and points of interest, thereby allowing BVI to navigate safely and confidently. Real-time assistance has also emerged as the key feature. Providing immediate feedback on the environment enables users to travel efficiently and safely, which leads to increased mobility and independence. 
Semantic mapping generates maps beyond geometric data. Such representations contain not only spatial information but also the semantic meanings of objects and features within the environment. This semantic understanding offers a deeper insight into the environment. This contextual awareness is particularly beneficial for enhancing navigation accuracy, as it enables navigation systems to make decisions based on semantic context, improving obstacle avoidance, path planning, and overall navigation efficiency. 

By employing robotic systems such as small robots, smart canes, and sensor-equipped suitcases, some studies have provided guidance, obstacle avoidance capabilities, and increased spatial awareness, thereby effectively providing independent navigation for visually impaired individuals.
Smartphone-based solutions harness the ubiquitous nature of smartphones that are equipped with cameras and sensors. These solutions offer navigation assistance by using widely available and familiar devices.
Both indoor and outdoor navigation capabilities offer a seamless transition between environments, ensuring that users receive consistent support, regardless of the scene in which they navigate.
Innovative localization and mapping algorithms enhance navigation efficiency and effectiveness through tailored modifications of existing SLAM frameworks or through the creation of novel solutions. Ultimately, these advancements have led to an improved overall experience for individuals with visual impairment.
Although these studies focused on different features and attributes, they all aimed to enhance mobility, independence, and overall quality of life for BVI people.

\input{texs/tables/t14_RQ4_mobility_improvement}

%
%
\subsubsection{Effectiveness of localization and mapping techniques in real-world scenarios}
The effectiveness of the localization and mapping techniques in real-world scenarios varies across studies. Tables \ref{tab:t_effectiveness_indoor_part_I}-\ref{tab:t_effectiveness_out_both} summarize these evaluations, highlighting key attributes such as the working area, localization and mapping accuracy level, robustness level, consideration of dynamic objects, and running time.
The robustness and accuracy levels reported in these tables are extracted from each paper's context; each rating reflects conditions specific to that paper and is not necessarily superior or inferior to the other approaches. Thus, these values are not comparable due to differing conditions across the papers.

Many studies, such as \cite{li2022sensing}, \cite{zhang2023research}, and \cite{hou2022knowledge}, have demonstrated high localization and mapping accuracy, particularly in indoor environments. These studies employed techniques such as ORB-SLAM2 and Cartographer to ensure reliable feature matching and adaptive navigation.

Robustness is another critical factor, with many systems proving to be resilient under various conditions. Studies such as \cite{zhou2022multi} and \cite{cheng2021hierarchical} reported high robustness owing to the integration of multiple sensors and multi-modal imaging. These systems can navigate complex environments and maintain accurate localization.

However, some studies highlighted some challenges. For instance, \cite{lu2021assistive} indicated that SLAM-based systems struggle with dynamic environments, leading to unstable navigation and orientation errors. Similarly, \cite{son2022wearable}pointed out issues with SLAM-relative poses in changing or occluded feature scenarios that affect navigation stability.

The running time is another essential consideration, with many studies emphasizing real-time performance. Systems such as those described in \cite{chen2021wearable} and \cite{slade2021multimodal} provide real-time performance, which is crucial for assistive navigation. However, some systems, such as those in \cite{liu2020indoor}, face longer computational times owing to their increased complexity, which can be a drawback in real-world applications.

Overall, the evaluation of localization and mapping techniques across different studies revealed a range of performance levels. High accuracy and robustness are common in controlled indoor environments, whereas dynamic and complex scenarios pose significant challenges. Insights from these evaluations are crucial for understanding the practical applicability and limitations of SLAM-based assistive systems for visually impaired individuals.

\input{texs/tables/t_effectiveness_indoor_part_I}
\input{texs/tables/t_effectiveness_indoor_part_II}
\input{texs/tables/t_effectiveness_out_both}

%
%
\subsubsection{User-based evaluations}
This section analyzes the user-based evaluations conducted to assess the satisfaction of the proposed SLAM-based assistive systems. By examining these evaluations, we gained insights into the real-world applicability and user acceptance of these technologies.
Several studies conducted user-based evaluations with actual participants to assess the effectiveness of and satisfaction with their proposed systems. These evaluations provided valuable insights into the usability and acceptance of assistive technologies. Tables \ref{tab:t_user_evaluation_user_based_partI} and \ref{tab:t_user_evaluation_user_based_partII} summarize studies that include user-based evaluations.

Three methods for assessing user satisfaction were identified: user studies, interviews, and surveys. Additionally, some studies involved only visually impaired participants, some involved only blindfolded users, and some included both groups to test their systems. Most studies used user studies as the primary evaluation method. Some studies also employed interviews or surveys after initial user studies to gather additional information on user satisfaction. The tables also show the experimental sites where the evaluations were conducted.

For example, \cite{ahmetovic2023sonification} involved nine BVI participants on a university campus to evaluate a sonification system and collect feedback on pleasantness, annoyance, precision, quickness, and overall appreciation. Similarly, \cite{li2022sensing} conducted evaluations with two BVI and three blindfolded participants in a laboratory setting, focusing on the task success rates, completion times, and feedback from verbal and haptic cues.

The study by \cite{song2023mixture} included five BVI and three blindfolded participants, achieving user satisfaction scores between six and nine out of ten. Another study by \cite{zhang2023research} evaluated their system with ten blindfolded participants, noting improvements in acceptance and trust levels.

\cite{hou2022knowledge} found moderate to high satisfaction among eight blindfolded participants, who found the system acceptable and useful for indoor navigation. In contrast, \cite{zhang2017indoor} highlighted that while users found the wayfinding function useful, they expressed discomfort owing to the weight of the device.
Overall, the user-based evaluations indicated that participants generally found the proposed systems beneficial for navigation, with varying levels of satisfaction based on the specific features and implementation of each system.

Some studies only conducted technical tests, without involving direct user feedback. These studies are summarized in Table \ref{tab:t_user_evaluation_technical_test}. For example, \cite{qiu2022egocentric} and \cite{chen2021wearable} focused on the technical performance of their systems and conducted tests in controlled environments but did not report user satisfaction.

The absence of user-based evaluations limits our understanding of how these systems perform in real-world scenarios and their acceptance among users. Future research should aim to incorporate comprehensive user studies to complement technical assessments and provide a more holistic view of a system’s effectiveness and usability.

\input{texs/tables/t_user_evaluation_user_based_partI}
\input{texs/tables/t_user_evaluation_user_based_partII}
\input{texs/tables/t_user_evaluation_technical_test}

\subsubsection{System prototype information}
To provide a comprehensive understanding of the assistive solutions proposed in the reviewed studies, we present the information regarding the system prototypes in Tables \ref{tab:t_prototype_information_wearable_partI}-\ref{tab:t_prototype_information_misc}. These tables include data on the functionalities, sensors used, computing resources, human-computer interaction (HCI) mechanisms, assistive tools, battery life, and whether the solutions are machine learning-based. Notably, the specifications in these tables cover the entire assistive system, and not just the localization and mapping components, as presented in Tables \ref{tab:core_technical_features_partI}-\ref{tab:core_technical_features_partIII}.

\input{texs/tables/t_prototype_information_wearable_partI}
\input{texs/tables/t_prototype_information_wearable_partII}
\input{texs/tables/t_prototype_information_handheld_partI}
\input{texs/tables/t_prototype_information_handheld_partII}
\input{texs/tables/t_prototype_information_misc}

Functionalities include the capabilities and features of the assistive system, such as navigation, object recognition, and obstacle avoidance. Sensors specify the types of sensors used in assistive devices such as cameras, LiDAR, and IMUs. Computing Resource indicates the hardware used for processing, including local devices, such as smartphones and laptops, as well as remote servers. HCI describes the interaction mechanisms used to provide feedback to the user, such as audio and haptic feedback. The assistive tool details the form factor of assistive devices, such as smart glasses, canes, and robot systems. Battery Life provides information on the operational duration of a device on a single charge. ML-based indicates whether the assistive solution incorporates machine learning algorithms.

Table \ref{tab:t_functionalities_classification} categorizes the papers based on the functionalities offered by assistive systems, highlighting the diverse capabilities ranging from basic navigation and obstacle avoidance to advanced features such as scene understanding and social networking. Table \ref{tab:t_hci_classification} classifies the papers based on the HCI mechanisms employed, showing the prevalence of audio feedback and the growing trend towards multimodal feedback incorporating haptic and tactile cues. Finally, Table \ref{tab:t_assistive_tool_classification} categorizes the studies based on the form factor of the assistive tool, revealing the diversity of approaches, including smartphone-based, wearable devices, handheld devices, and robotic systems.

\input{texs/tables/t_functionalities_classification}
\input{texs/tables/t_hci_classification}
\input{texs/tables/t_assistive_tool_classification}

\paragraph{Functionalities}
The analysis of the data in the tables shows the diverse range of approaches and technologies used to create assistive systems for visually impaired navigation. Most systems focus on navigation and obstacle avoidance, but many also include advanced features such as scene understanding and social networking. The use of sensors is diverse, with RGB-D cameras being the most commonly used because of their capability to capture both color and depth information, especially for the localization and mapping components of assistive systems.

\paragraph{HCI}
The HCI mechanisms vary, with audio feedback being the most commonly used method. Several systems also use haptic feedback and a few incorporate visual hints for users with partial vision. These feedback mechanisms are essential for real-time navigation assistance and for enhancing user experience.
Several studies used multimodal feedback for real-time navigation assistance, as indicated by the HCI column. These include combinations of audio, haptic, and grounded kinesthetics to enhance user experience and provide comprehensive navigation aids. This multimodal approach ensures that users receive complementary information, thereby enhancing the robustness and reliability of assistive systems.

\paragraph{Assistive tool}
The form factors of assistive tools vary among the studies. Wearable devices such as smart glasses and helmets are designed to be worn on the body and provide hands-free assistance. Handheld devices, such as smart canes, are traditional mobility aids enhanced by modern technology. These smart canes include sensors to detect obstacles and provide real-time feedback through vibrotactiles or steering. This approach leverages the familiarity and comfort of using a cane, while adding significant technological advancements to aid navigation and spatial awareness. Some prototypes incorporate both wearable and handheld components; for example, in the study by \cite{chen2017ccny}, a Google Tango device was mounted on the user's chest while the user held a smart cane. These prototypes are categorized as handheld devices because the users’ hands are occupied. 
Robotic systems represent another innovative factor. These can range from small mobile robots in the shape of a suitcase that guides users through complex environments to more substantial ride-on systems such as autonomous wheelchairs or scooters.

\paragraph{Battery life}
The battery life is a critical factor in assistive navigation systems. These systems must be reliable to ensure continuous assistance without frequent recharging interruptions. 
The battery lives of the proposed solutions varied across the reviewed studies. Some systems, such as those described by \cite{li2022sensing},  have reported a long battery life, which ensures that the devices remain functional during extended use.
However, not all studies provide detailed information on battery life. This lack of information can be a concern, as it leaves uncertainty regarding the reliability of the device in real-world scenarios.
Additionally, some devices, such as those incorporating high-performance processors or multiple sensors, may face challenges in maintaining a long battery life owing to their higher power consumption. Systems such as those described by \cite{son2022wearable}, which use advanced components such as the Nvidia Jetson AGX Xavier, may offer robust functionality, but require careful management of power resources to ensure adequate battery life.

\paragraph{Machine-learning approaches}
Many systems leverage machine learning for functionalities, such as object detection, scene understanding, and localization. Algorithms such as YOLO, Faster R-CNN, and various deep neural networks are commonly employed. These machine learning-based solutions enhance the accuracy and efficiency of the assistive systems.
Table \ref{tab:ml_categorization} categorizes the machine learning approaches used in assistive devices, along with their references. This categorization illustrates the diversity of machine-learning techniques applied to improve the functionalities of assistive systems for BVI navigation.

\input{texs/tables/t_ml_categorization}
%
%
\section{Future opportunities} 
\label{future_opportunities}
An effective navigation system for BVI people needs to meet mobility metrics, such as decreasing navigation time, decreasing navigation distance, decreasing contact with the environment, and increasing walking speed. These systems must be highly accurate and efficient in complex situations, such as crowded places and changing light and weather conditions.  At the same time, assistive aids should be comfortable, easy to use, unobtrusive, cost-effective, lightweight, and reduce cognitive load. 

In this review, we examined publications that employed SLAM techniques in their navigation approaches. One of the distinct advantages of SLAM is its applicability in diverse locations without the need for pre-built maps or additional infrastructure such as Bluetooth beacons or RFID tags. However, there is still room for improvement in various aspects of these systems, including their ability to handle complex scenarios, provide accurate obstacle information, and seamlessly transition between indoor and outdoor environments.

There is a notable lack of focus on adapting SLAM to challenging situations especially dynamic environments, specifically for visually impaired navigation. Studies in this field typically employ SLAM as a pre-existing tool or use it without any significant adaptation to effectively handle challenges. Researchers in this area can draw inspiration from recent advancements in robotics to address this gap. By leveraging cutting-edge techniques from robotic navigation, SLAM systems that are more robust and suitable for assisting visually impaired individuals in real-world scenarios can be developed.

This section discusses the open problems and research directions identified during the SLR.

\textbf{Challenge scenarios and real-world studies}
Navigating crowded environments remains a significant challenge for the visually impaired and studies addressing this issue are limited. Evaluations are often conducted in controlled settings rather than real-world scenarios. This is problematic because controlled environments may not accurately reflect the dynamic and unpredictable nature of real-world crowded spaces, where visually impaired individuals face numerous obstacles and safety risks. Future research should focus on developing and testing solutions in high-traffic public places such as train stations and shopping malls. This is crucial to ensure that assistive navigation systems can effectively handle the complexities of real-world crowded environments, including the presence of numerous dynamic objects, varying crowd densities, and unpredictable pedestrian behaviors. 

Furthermore, addressing challenging conditions such as changes in illumination, low-light scenarios, high-speed dynamic objects, and complex backgrounds can enhance the robustness and versatility of navigation systems. These challenging conditions are common in real-world scenarios and can significantly impact the performance and reliability of SLAM-based navigation systems. By addressing these challenges, researchers can develop more robust and adaptable solutions that can function effectively in diverse and demanding environments. 

Techniques such as image enhancement for ORB points and LSD line feature recovery used in agricultural environments \cite{islam2023agri} can be adapted for visually impaired navigation.
Adapting these techniques from other domains can accelerate the development of more effective solutions for visually impaired navigation, leveraging existing knowledge and expertise to address the specific challenges faced by this user group.

\textbf{Long-term navigation}
The development of solutions that are effective over extended navigation periods is critical to achieve autonomous navigation. These solutions must ensure accurate mapping and localization even when the maps are updated over a longer navigation duration. 
This is important because environments are not static; they change over time. Obstacles may appear or disappear, and landmarks may be altered. A SLAM-based navigation system that cannot adapt to these changes will become increasingly inaccurate and unreliable over time, potentially leading to dangerous situations for visually impaired users.

To address this challenge, researchers can leverage solutions proposed in robotics, such as those presented in \cite{deng2023long}, which introduced a novel long-term SLAM system with map prediction and dynamic removal, thereby allowing wheelchair robots to maintain precise navigation capabilities over extended periods.

Future research should focus on the development of robust algorithms for continuous map updates and maintenance, including strategies for handling environmental changes over time.
These strategies are essential to ensure that the navigation system can maintain its accuracy and reliability over extended periods, even in the face of environmental changes. By continuously updating and refining the map, the system can provide visually impaired users with up-to-date and relevant information about their surroundings, enabling them to navigate safely and confidently.

\textbf{Deep learning integration}
The integration of deep learning with the SLAM algorithms for BVI navigation requires further investigation. Deep learning offers a versatile approach for enhancing various aspects of SLAM such as precise pose estimation under challenging conditions, relocalization, and loop-closure detection. 
This is important because deep learning can potentially improve the accuracy and robustness of SLAM in complex and dynamic environments, where traditional SLAM algorithms may struggle. For instance, deep learning can be used to improve feature detection and matching in low-light conditions or to predict and adapt to changes in the environment.
Despite challenges, such as the need for large, accurately labeled datasets, the black-box nature of deep-learning models, and the computational intensity, the association between deep learning and SLAM holds promise for advancing navigation solutions for the visually impaired, particularly in challenging scenarios.
The potential benefits of deep learning for SLAM are substantial, and overcoming these challenges could lead to significant advancements in assistive navigation technology.

Future research should focus on developing more efficient deep learning models that can operate effectively with limited computational resources and real-time constraints. 
This is crucial because visually impaired individuals need real-time feedback and guidance to navigate safely and effectively. Deep learning models that are computationally intensive or require powerful hardware may not be practical for real-world use.
Additionally, creating large-scale, accurately labeled datasets tailored for BVI navigation is crucial for training robust models. 
The lack of such datasets is a major obstacle to the development of effective deep-learning-based SLAM systems for visually impaired navigation.

Addressing the interpretability of deep learning models can also enhance the trust and transparency in these systems. 
Collaboration among machine learning experts, roboticists, and vision scientists can drive the development of innovative algorithms that leverage deep learning to enhance the reliability and accuracy of SLAM-based navigation aids for the visually impaired.
This collaboration is essential to bring together the diverse expertise needed to develop effective and practical solutions.

\textbf{Indoor and outdoor navigation integration}
Seamless transitions between indoor and outdoor environments are crucial for enhancing the independence and mobility of BVI individuals. However, most studies in our SLR have focused primarily on indoor environments. 
This limitation arises because indoor environments are often more structured and predictable than outdoor environments, making them easier to map and navigate using SLAM. Outdoor environments, on the other hand, present challenges such as varying lighting conditions, weather changes, and a wider range of obstacles.

Future research should aim to develop solutions that provide unified and consistent navigation experience in both indoor and outdoor settings.
This is important because visually impaired individuals need to be able to navigate seamlessly between different environments in their daily lives. A navigation system that only works indoors or outdoors would be of limited use.
Researchers should explore the integration of robust sensor fusion techniques and adaptive algorithms capable of handling the different conditions and challenges of these environments. 
Sensor fusion can combine data from multiple sensors, such as cameras, LiDAR, and IMUs, to provide a more comprehensive and accurate understanding of the environment. Adaptive algorithms can adjust the SLAM system's parameters in real time to account for changes in lighting, weather, and other environmental factors.

\textbf{Obstacle detection}
Achieving detailed knowledge of obstacles and their characteristics is essential for BVI people. Although some studies included in the SLR addressed obstacle detection, the depth and accuracy of the obstacle information provided may still be limited. 
This limitation stems from the fact that traditional obstacle detection methods often focus on identifying the presence and location of obstacles but may not provide detailed information about their shape, size, or material, which is crucial for visually impaired individuals to make informed decisions during navigation.

To address the need for more detailed and accurate obstacle information, future research should focus on advancing SLAM algorithms to deliver context-aware obstacle detection. This involves integrating semantic understanding with precise spatial measurements, allowing the system to identify and interpret the nature and significance of obstacles accurately.
By incorporating semantic understanding, SLAM systems can differentiate between different types of obstacles, such as curbs, stairs, or low-hanging branches, and provide more relevant and actionable information to the user. Precise spatial measurements are essential for accurately estimating the distance, size, and shape of obstacles, enabling visually impaired individuals to navigate safely around them.

In addition, it is crucial to develop algorithms that can learn and adapt to various obstacle types and scenarios. 
Drawing inspiration from the approaches used in robotics and autonomous drones, such as the real-time metric-semantic SLAM demonstrated by \cite{rosinol2020kimera}, can provide valuable insights. 
These approaches have demonstrated the feasibility and effectiveness of integrating semantic understanding with precise spatial measurements in real-time SLAM systems.
Therefore, future research should prioritize improving both the depth and accuracy of obstacle information, while ensuring robust real-time performance and adaptability to various real-world conditions.

\textbf{Semantic information integration}
Integrating semantic information into SLAM algorithms can significantly enhance the performance and robustness of navigation systems for BVI individuals. This information can be used to refine the mapping and localization processes and enhance the overall reliability of navigation in complex environments. For instance, semantic information aids in rejecting outliers during loop-closure detection, which is a crucial SLAM step that identifies and matches previously visited locations.
To advance this area, future research should focus on developing advanced techniques for semantic data extraction and integration within the SLAM frameworks. 
This is necessary because current methods for semantic data extraction and integration may not be efficient enough for real-time SLAM in complex environments. By developing more advanced techniques, researchers can improve the quality and reliability of semantic information used in SLAM, leading to better navigation performance.

Researchers should also explore methods to ensure real-time performance while maintaining the accuracy and detail of the semantic information.
Real-time performance is crucial for providing visually impaired users with timely and relevant feedback during navigation. However, processing and integrating semantic information is computationally expensive. Therefore, it is important to develop methods that can balance real-time performance with the accuracy and detail of semantic information.

\textbf{Realistic dataset creation}
Another crucial area for future research is the development of realistic and comprehensive datasets tailored for BVI navigation. Although some datasets exist for various SLAM applications, they often do not capture the unique challenges faced by the visually impaired, such as the need to navigate crowded spaces and avoid obstacles at different heights. Future research should focus on creating large-scale, diverse datasets that include various indoor and outdoor settings, different lighting conditions, and dynamic elements. This is important because the lack of such datasets hinders the development and evaluation of SLAM algorithms that are specifically designed for BVI navigation. By creating datasets that reflect real-world challenges faced by visually impaired individuals, researchers can develop more effective and reliable navigation solutions.

\textbf{Computing resources and battery life}
The development and deployment of SLAM-based navigation systems for the visually impaired face significant challenges related to computing resources and battery life. SLAM algorithms, particularly when integrated with deep learning models, often demand substantial computing power that can quickly drain battery life and generate heat in mobile devices. This is a critical issue because visually impaired individuals need portable and comfortable navigation aids that can operate for extended periods without overheating or frequent recharging. 

Additionally, intensive computations and continuous sensor usage drain the battery life quickly, limiting the usability of the system in real-world daily scenarios. This limitation can significantly hinder the adoption and effectiveness of the SLAM-based navigation systems. Future research should focus on optimizing SLAM algorithms and deep learning models for low-power devices without compromising the accuracy or real-time performance. This is essential for developing energy-efficient solutions that can operate on portable devices with limited battery capacity, ensuring that visually impaired individuals can rely on these systems for extended periods without interruption. Exploring edge computing solutions, developing more efficient neural network architectures, and enhancing battery management techniques could help to address these challenges. These approaches can collectively contribute to reducing the computational burden and power consumption of SLAM-based navigation systems, making them more practical and sustainable for real-world use.

\textbf{Human-computer interaction for SLAM-based assistive devices}
An important area for future research is to improve the human-computer interaction (HCI) aspects of SLAM-based assistive devices for BVI individuals. Although SLAM techniques have shown great potential in gathering and processing environmental information, effectively communicating this information to BVI users remains a significant challenge. Future research should focus on developing intuitive and non-intrusive methods to convey complex spatial data and navigational instructions to BVI users. This includes:
\begin{itemize}
    \item Multi-modal feedback systems: Exploring combinations of audio, haptic, and other non-visual feedback methods to provide rich contextual information without overwhelming the user.
    \item Adaptive interfaces: Developing interfaces that can adjust the level and type of information provided based on the user's preferences, familiarity with the environment, and the current situation.
    \item Natural language processing: Improving the ability of systems to understand and respond to natural language queries, allowing for a more intuitive interaction between the user and device.
    \item Cognitive load optimization: Investigating ways to balance the provision of detailed environmental information, ensuring that users receive the necessary guidance without cognitive overload.
    \item Real-time situational awareness: Developing methods to effectively communicate dynamic elements of the environment, such as moving obstacles or changing traffic conditions in real-time.
\end{itemize}

Addressing these HCI challenges will be crucial in translating the technical capabilities of SLAM into practical, user-friendly assistive devices that can significantly enhance the mobility and independence of BVI individuals. Future research in this area should involve close collaboration with BVI users to ensure that the developed interfaces meet their needs and preferences.

\textbf{Product development and collaboration}
Notably, all reviewed approaches were prototypes in the early stages of research and are not yet practical. This might be due to the absence of a unified community or group dedicated to solving the BVI navigation challenges. Much of the work in this domain has been conducted by academic groups or small companies that often fail to produce feasible final products. This underscores a significant future opportunity to develop collaboration and to bridge the gap between research and practical implementation.

Additionally, efforts should be made to develop standardized evaluation metrics and protocols to ensure that the developed systems meet real-world needs and can be effectively transitioned from prototypes to market-ready solutions. 
Standardized evaluation metrics and protocols are essential to ensure that assistive navigation systems are evaluated consistently and objectively. This can help identify the strengths and weaknesses of different approaches and guide the development of more effective solutions.
Encouraging partnerships with technology companies can also accelerate the commercialization process. These partnerships provide the necessary support to bring innovative solutions to the market.

In conclusion, the future of SLAM for the visually impaired navigation is promising. Continued research efforts have the potential to develop SLAM algorithms tailored for BVI navigation, empowering visually impaired individuals with a safe and independent means of navigating their surroundings.
\section{Conclusion} 
\label{conclusion}
This study presents a systematic literature review of recent studies on SLAM-based solutions for BVI navigation. Excluding papers published before 2017, this review focused on the latest advancements, innovations, and considerations, resulting in a more relevant and comprehensive understanding of the current state of research. The insights provided by this systematic literature review are intended to guide researchers in the academic and research communities. They inform the existing gaps and future opportunities to address the challenges faced by SLAM-based assistive solutions. 

Relevant data were extracted from 54 selected studies that adhered to the SLR selection criteria to address the research questions. By analyzing the selected papers based on their SLAM techniques, we observed that the majority of the studies utilized visual SLAM techniques, such as ORB-SLAM3, owing to their advantages for visual sensors. 

Several studies have introduced novel strategies for addressing localization and mapping challenges tailored to the specific requirements of their research, whereas certain studies have employed existing spatial tracking frameworks to develop navigation solutions.
We also investigated the advantages and limitations of the SLAM techniques, as highlighted in the studies under review. Notably, most studies have leveraged accurate localization features of SLAM. 

We investigated the challenging scenarios encountered by SLAM-based navigation systems, which have been addressed in the literature.
Additionally, we discussed practical challenges and considerations that affect the usability and adoption of these systems.
Furthermore, we analyzed how the proposed SLAM-based solutions improve the mobility and navigation of visually impaired individuals. We evaluated the effectiveness of these solutions in real-world scenarios and assessed the user satisfaction to understand their practical impact on BVI mobility.
Finally, we identified gaps, opportunities, and areas of interest that could be explored further in future research, such as addressing challenges in crowded environments, improving real-world applicability, integrating deep learning, and ensuring long-term navigation effectiveness in SLAM-based solutions for visually impaired navigation.

Given the widespread application of SLAM in robotic, autonomous drones, and auto-driving car navigation, these techniques can be adapted to ensure safe and independent BVI navigation. This is particularly important in dynamic and challenging environments, including those with varying lighting conditions where research opportunities remain abundant. The potential of integrating these techniques into the navigation of visually impaired individuals continues to be an open and promising avenue.

\section*{Acknowledgment}
We would like to extend our sincere gratitude to Giovanni Cioffi, whose insightful feedback and thorough review significantly contributed to the refinement of this manuscript.

\begin{IEEEbiography}[{\includegraphics[width=1in,height=1.25in,clip,keepaspectratio]{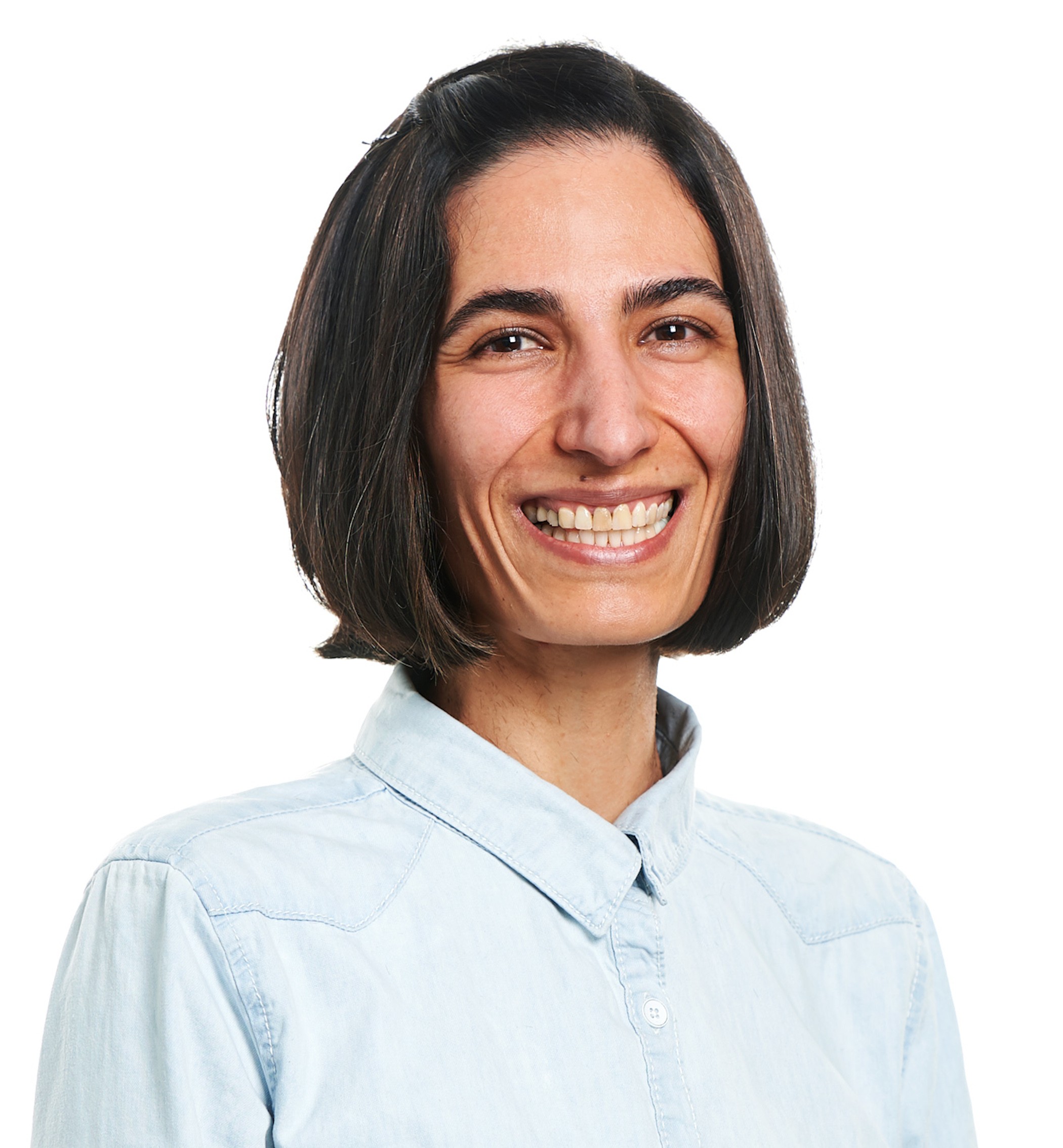}}]{Marziyeh Bamdad} is currently pursuing her Ph.D. in computer science at the University of Zurich, supervised by Prof. Davide Scaramuzza, and serves as a research assistant at the Zurich University of Applied Sciences in Switzerland. Her Ph.D. research is dedicated to developing innovative solutions for visually impaired navigation, harnessing the potential of Visual Simultaneous Localization and Mapping (Visual SLAM) technologies.
\end{IEEEbiography}
\begin{IEEEbiography}[{\includegraphics[width=1in,height=1.25in,clip,keepaspectratio]{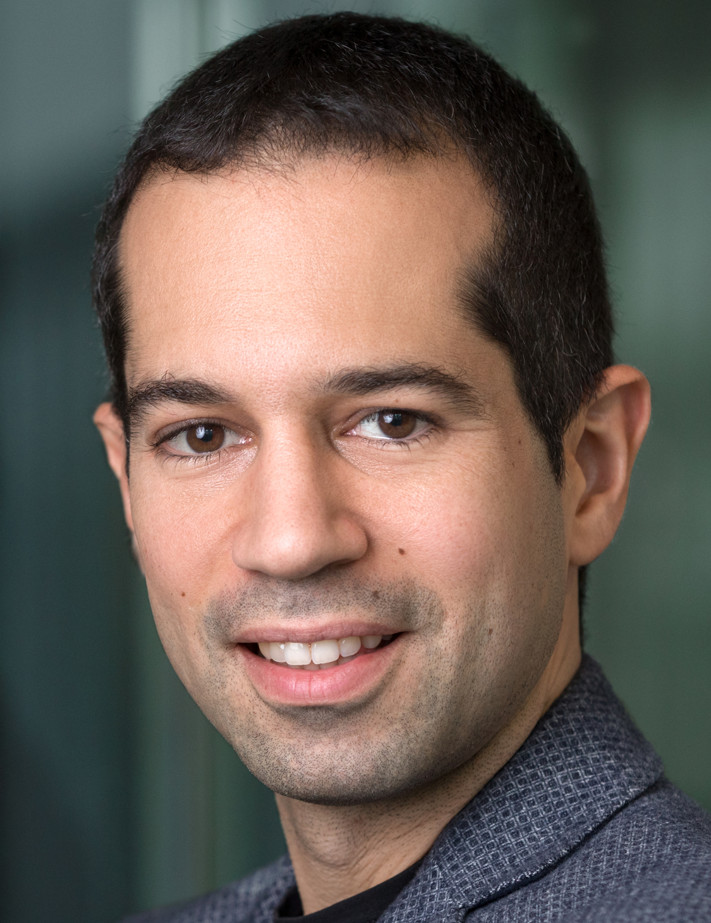}}]{Davide Scaramuzza} is a Professor of Robotics and Perception at the University of Zurich. He did his Ph.D. at ETH Zurich, a postdoc at the University of Pennsylvania, and was a visiting
professor at Stanford University. His research focuses on autonomous, agile microdrone navigation using standard and event-based cameras. He pioneered autonomous, vision-based navigation of drones, which inspired the navigation algorithm of the NASA Mars helicopter and many drone companies. He contributed significantly to visual-inertial state estimation, vision-based agile navigation of micro-drones, and low-latency, robust perception with event cameras, which were transferred to many products, from drones to automobiles, cameras, AR/VR headsets, and mobile devices. In 2022, his team demonstrated that an AI-controlled, vision-based drone could outperform the world champions of drone racing, a result that was published in Nature. He is a consultant for the United Nations on disaster response and disarmament. He has won many awards, including an IEEE Technical Field Award, the IEEE Robotics and Automation Society Early Career Award, a European Research Council Consolidator Grant, a Google Research Award, two NASA TechBrief Awards, and many paper awards.
In 2015, he co-founded Zurich-Eye, today Meta Zurich, which developed the world-leading virtual-reality headset Meta Quest. In 2020, he co-founded SUIND, which builds autonomous drones for precision agriculture. Many aspects of his research have been featured in the media, such as The New York Times, The Economist, and Forbes.
\end{IEEEbiography}
\begin{IEEEbiography}[{\includegraphics[width=1in,height=1.25in,clip, keepaspectratio]{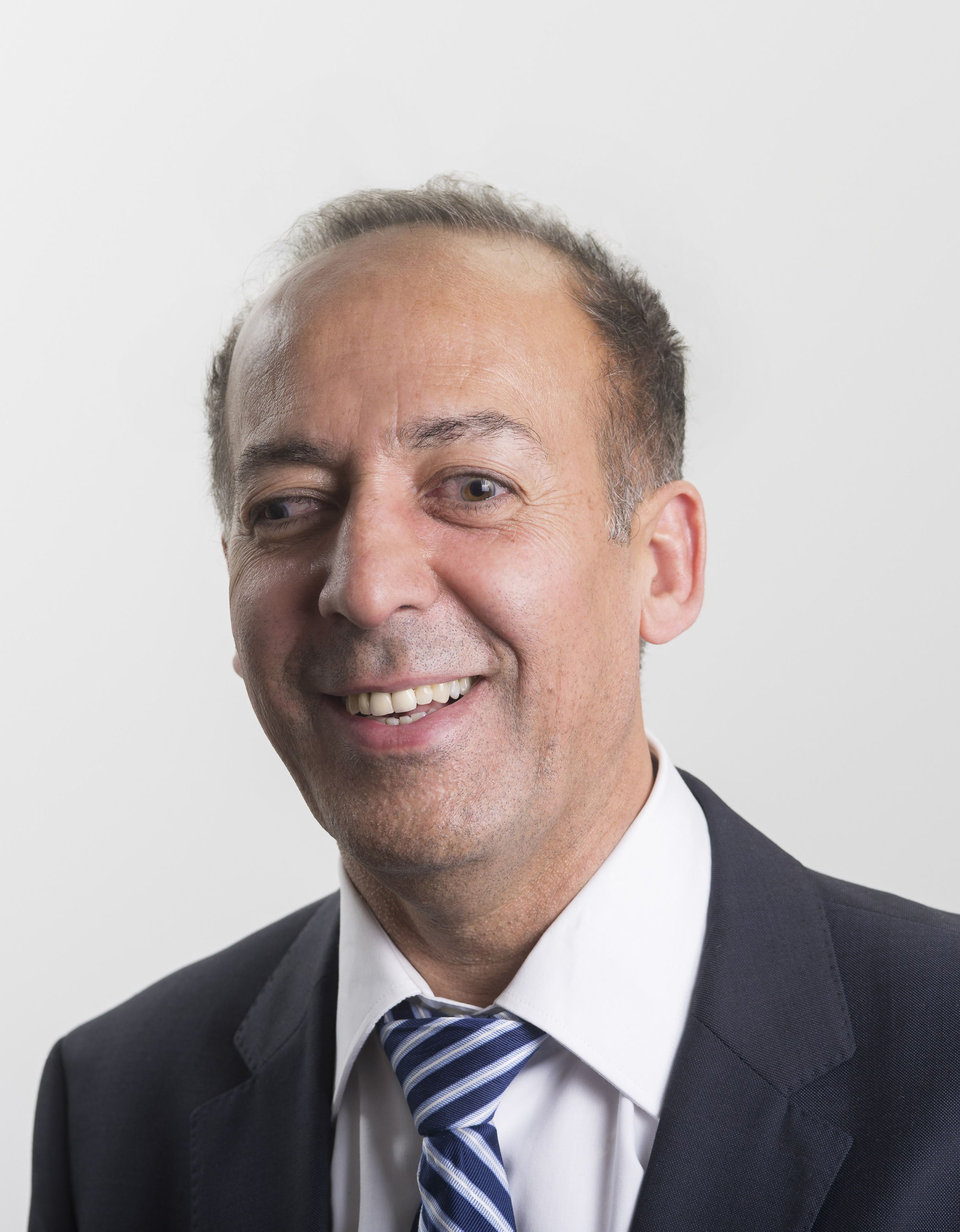}}]{Alireza Darvishy} is professor for ICT Accessibility and head of the ICT Accessibility Lab at Zurich University of Applied Sciences in Switzerland. He serves an independent reviewer for European research projects such as the Active Assisted Living (AAL) program, and is principle investigator of the “Accessible Scientific PDFs for All” project, funded by the Swiss National Science Foundation.

\end{IEEEbiography}

\EOD

\end{document}